\documentclass[11pt]{article}
\usepackage{bbm}

\usepackage[preprint]{acl}

\usepackage{times}
\usepackage{latexsym}

\usepackage[T1]{fontenc}

\usepackage[utf8]{inputenc}

\usepackage{microtype}

\usepackage{inconsolata}

\usepackage{graphicx}
\usepackage{amsmath}
\usepackage{booktabs}
\usepackage{multirow}
\usepackage[dvipsnames]{xcolor}
\usepackage[most]{tcolorbox}
\usepackage{float}
\usepackage{subcaption}
\usepackage[table,dvipsnames]{xcolor}
\usepackage{multirow}
\usepackage{booktabs}

\definecolor{LawBG}{HTML}{E8D4E8}         
\definecolor{FinanceBG}{HTML}{CFDFF6}     
\definecolor{HealthcareBG}{HTML}{C9ECDF}  
\definecolor{OfficeBG}{HTML}{EBDDC3}      
\definecolor{MathBG}{HTML}{FBE1C2}        

\tcbuselibrary{listings,breakable}
\newtcblisting{skillbox}{
  colback=gray!3,
  colframe=gray!50,
  sharp corners,
  boxrule=0.5pt,
  left=6pt,
  right=6pt,
  top=6pt,
  bottom=6pt,
  breakable,
  listing only,
  listing options={
    basicstyle=\ttfamily\scriptsize,
    breaklines=true,
    columns=fullflexible,
    keepspaces=true,
    showstringspaces=false
  }
}
\title{
ContinualSkillBench: Can LLM Agents Truly Evolve Their Capabilities?
\thanks{Source code: \url{https://github.com/gtynnn060110-hash/continual-skill-bench-final}.}
}

\author{
Tianyi Guan\textsuperscript{\rm 1}\thanks{\ \ Equal Contribution.} \quad
Yiding Wang\textsuperscript{\rm 1}\footnotemark[2] \quad
Haotong Yang\textsuperscript{\rm 1} \quad
Siyuan Cao\textsuperscript{\rm 1} \quad
Shirui Liu\textsuperscript{\rm 1} \quad
Yi Hu\textsuperscript{\rm 1} \\
\textbf{Jiaqi Li}\textsuperscript{\rm 2}\thanks{\ \ Corresponding Authors.} \quad
\textbf{Muhan Zhang}\textsuperscript{\rm 1, \rm 2}\footnotemark[3] \\
\textsuperscript{\rm 1}Institute for Artificial Intelligence, Peking University \\
\textsuperscript{\rm 2}Beijing Institute for General Artificial Intelligence
}

\begin{document}
\maketitle
\begin{abstract}

Modern agent frameworks equip large language models with external skill libraries to solve complex tasks. However, it remains unclear whether these systems can effectively evolve their skills and whether the resulting skills improve task-solving capabilities. To bridge this gap, we introduce \textbf{\textsc{ContinualSkillBench}}, a dynamic evaluation framework for in-context continual skill learning. It covers five domains, each with 100 tasks ordered by difficulty and opportunities for skill transfer. Our experiments show that sequential execution generally improves performance, but the gains vary substantially across models and domains. Moreover, across two models and three domains, retaining context and feedback without skill maintenance yields nearly the same mean performance as maintaining an explicit skill library. Explicit skills nevertheless provide selective benefits for tasks requiring reusable procedures or precise outputs. We further find that the weaker model (GPT-4o) accumulates more skills but reuses them less often, suggesting difficulty consolidating task-specific experience into reusable procedures. These findings show that current in-context skill evolution mechanisms can support continual adaptation, but still struggle to consistently consolidate experience into robust and transferable skills.
\end{abstract}

\section{Introduction}

As LLMs increasingly serve as agents for complex real-world tasks, they often require domain-specific knowledge and procedures beyond their pretrained behavior. ``Agent skills'' ---reusable instructions and workflows typically stored as structured documents ---provide such capabilities \citep{schick2023toolformerlanguagemodelsteach,shen2023hugginggptsolvingaitasks} and are now supported by agent platforms including Claude Code \citep{anthropic2025claudecode} and Codex \citep{openai2025codexcli}.

Recent work has shown that given well-written skills from human experts, agents perform significantly better on corresponding tasks \citep{li2026skillsbenchbenchmarkingagentskills}. However, it is fundamentally difficult and expensive to manually build a comprehensive skill library for agents. In more realistic settings, where users only provide task descriptions and feedback through a task sequence, a critical question is whether agents can autonomously synthesize and evolve their own skills from these interactions. Although some works explore this direction \citep{clbench2026,zhong2026skilllearnbenchbenchmarkingcontinuallearning}, we still lack a systematic evaluation of their capability limits. Can experience from earlier tasks improve performance on later tasks? Can agents consolidate such experience into reusable skills, rather than merely adapting to recent feedback or output conventions? And does maintaining an explicit, evolving skill repository provide benefits beyond retaining prior context and feedback?

We study this setting through \textsc{ContinualSkillBench}, a comprehensive evaluation framework for agent skill evolution. Unlike previous evaluations that test agents on isolated tasks with fixed skill documents, \textsc{ContinualSkillBench} is explicitly designed to assess an agent's capacity for continuous feedback learning and to evaluate whether the dynamically evolved skills lead to capability expansion. Specifically, our benchmark includes five representative domains (e.g., Law, Healthcare, and Finance). Within each domain, we curate 100 interconnected subtasks anchored to three core skills. Rather than presenting these tasks at random, we strategically order them by varying levels of difficulty and underlying skill dependencies. This logical progression creates an environment in which downstream objectives implicitly rely on the skills practiced in earlier steps, thereby providing a realistic and expansive space for agents to engage in continual learning.


We evaluate multiple model--harness configurations and find that sequential execution improves in 14 of 15 model--domain combinations. However, pure ICL and explicit skill maintenance perform nearly identically on average across the two models and three domains in our ablation, indicating that retained context and feedback account for much of the observed gain. Their strengths differ by task type: explicit skills better support reusable procedures and rigid output requirements, whereas full-context ICL is often stronger on open-ended tasks. Repository analysis further reveals that accumulating more skills does not ensure effective learning: weaker agents produce larger but less frequently reused and lower-quality repositories. Overall, current agents can adapt through sequential interaction but still struggle to consolidate experience into transferable skills.

Our main contributions are as follows:
\begin{itemize}
    \item We introduce \textbf{\textsc{ContinualSkillBench}}, a dynamic evaluation framework designed to assess the continual skill evolution and in-context learning capabilities of LLM agents.

    \item We construct ordered task sequences across five domains, with each domain containing 100 interconnected subtasks organized by increasing difficulty and opportunities for cross-task skill
    reuse.

    \item We benchmark multiple model-harness configurations and compare explicit skill maintenance with independent and in-context learning baselines, revealing the benefits and limitations of current sequential agent frameworks.
\end{itemize}

\begin{figure*}[t]
  \centering 
  \vspace{-10pt}
  \includegraphics[width=\textwidth]{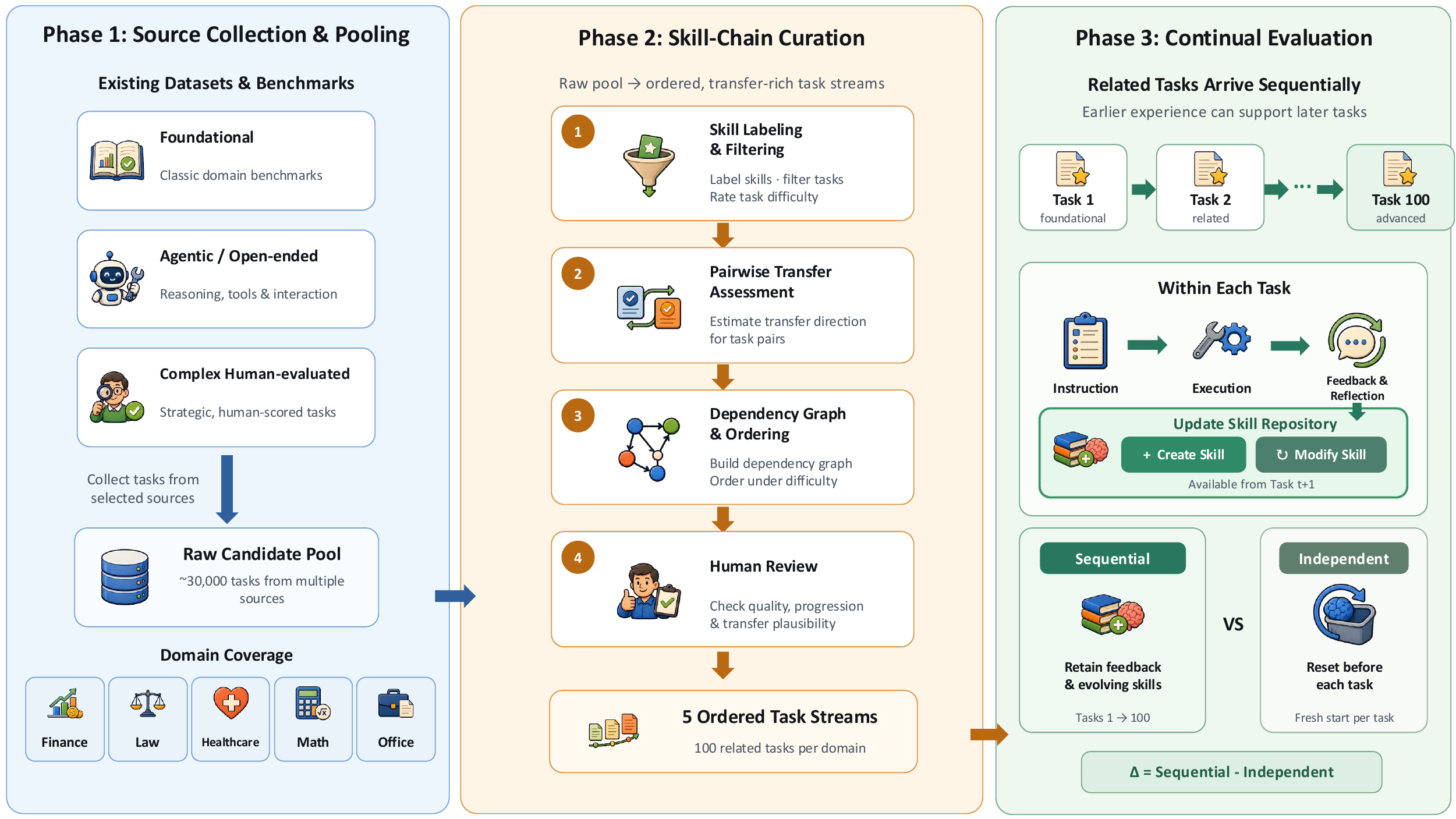}
  \caption{Overview of the \textsc{ContinualSkillBench} pipeline. Phase 1 collects approximately 30,000 tasks from five domains. Phase 2 uses LLM-assisted filtering, skill-dependency analysis, task ordering, and human review to construct five sequential task streams. Phase 3 evaluates agents through instruction, execution, and reflection, comparing Sequential and Independent execution with their performance difference reported as $\Delta$.}
  \label{fig:main}
  \vspace{-0.5cm}
\end{figure*}

\section{Related Work}

\textbf{In-Context Skill Evolution for Agents}.
To execute complex, domain-specific tasks, LLM agents increasingly rely on external tools or structured \textbf{skills} \citep{schick2023toolformerlanguagemodelsteach, shen2023hugginggptsolvingaitasks, xu2026agentskillslargelanguage}. However, most current approaches assume a static skill library, severely limiting adaptability in dynamic environments. Furthermore, adapting models to new sequential tasks via standard parametric updates (e.g., fine-tuning) typically leads to catastrophic forgetting \citep{Kirkpatrick_2017}. To mitigate this, recent research has explored \textit{in-context} continual learning \citep{wu2024continuallearninglargelanguage, raparthy2023generalizationnewsequentialdecision}. By leveraging extended context windows, agents can achieve lifelong learning—dynamically updating and retrieving a growing library of skills over sequential tasks—without fine-tuning their parameters.

\vspace{\baselineskip}
\noindent \textbf{Benchmarks for Agent Skills and Continual Learning}.
Most general-purpose agent benchmarks evaluate task completion under a fixed set of tools, instructions, or skills~\citep{mialon2023gaiabenchmarkgeneralai,zhang2026clawbenchaiagentscomplete,ye2026clawevaltrustworthyevaluationautonomous}. SkillsBench evaluates the utility of externally provided skills through paired no-skill and curated-skill conditions, but does not study how a skill library evolves over a sequence of interactions ~\citep{li2026skillsbenchbenchmarkingagentskills}. Recent benchmarks have begun to evaluate skill learning from experience. SkillLearnBench~\citep{zhong2026skilllearnbenchbenchmarkingcontinuallearning} studies several methods for generating and refining skills from seed instances, and evaluates the resulting skills through skill quality, execution behavior, and performance on additional instances of the same task. CL-Bench~\citep{dou2026cl} takes a broader view of online learning by constructing stateful task sequences with shared latent structure and comparing different in-context and memory-based agent architectures. SkillCraft~\citep{chen2026skillcraft} instead focuses on the acquisition and reuse of executable tool compositions across tasks. \textsc{ContinualSkillBench} complements these efforts by studying skill evolution over long, heterogeneous task streams within broad domains. Each stream contains tasks from multiple sources and evaluation formats, connected by recurring core skills rather than a single repeated task or a fixed execution workflow. This setting allows us to examine whether experience acquired from one task can be consolidated into structured skills and reused on different downstream tasks, as well as how this ability varies across domains and models.

\section{\textsc{ContinualSkillBench}}

\begin{figure*}[t]
    \centering 
    \vspace{-10pt}\includegraphics[width=0.8\textwidth]{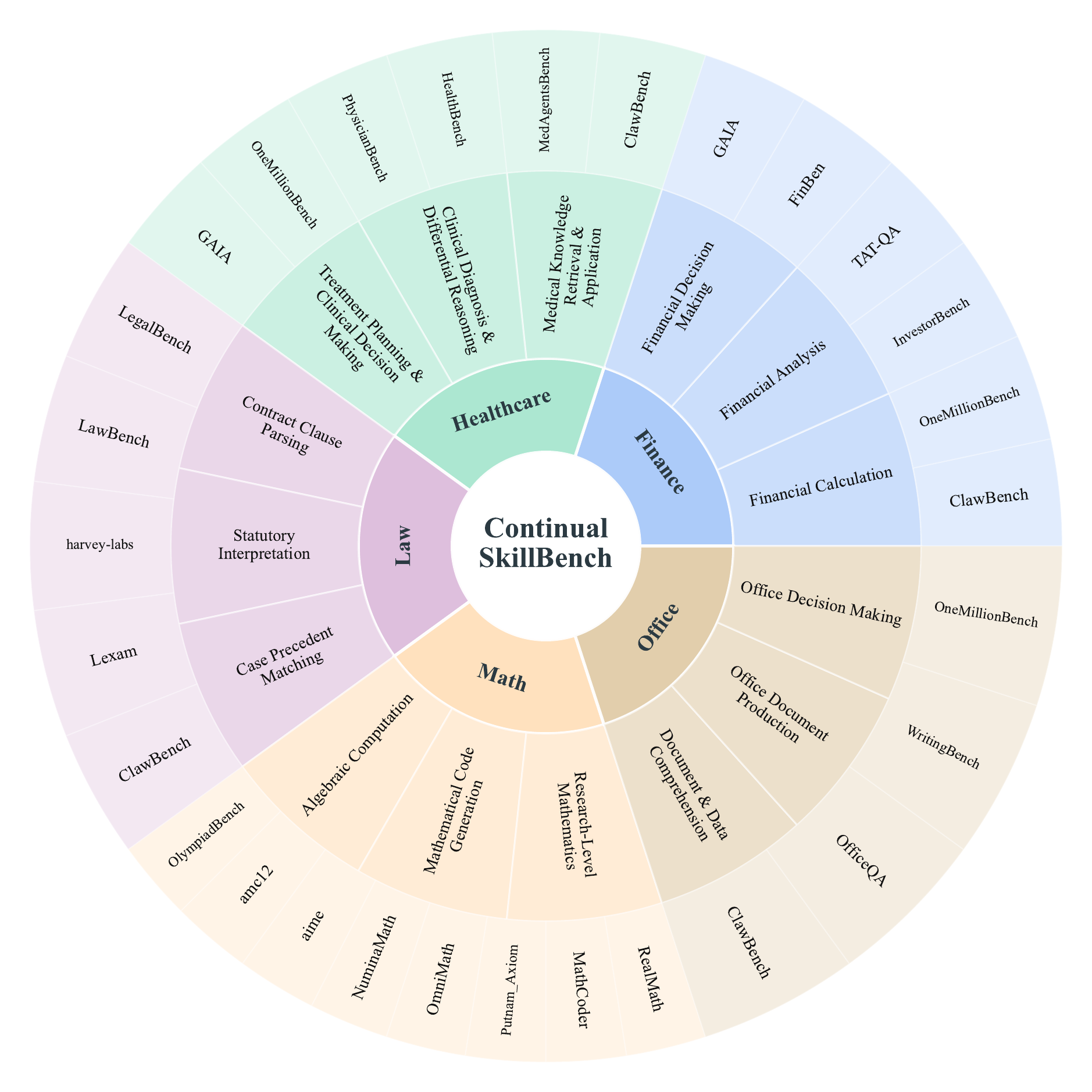}
    \caption{Taxonomy of \textsc{ContinualSkillBench}. The inner ring shows the five domains, the middle summarizes three macro capability tracks within each domain, and the outer lists the source datasets and benchmarks.}
    \label{fig:bench-composition}
    \vspace{-0.5cm}
\end{figure*}


To address the gap, we introduce \textsc{ContinualSkillBench} to evaluate agents' continual skill learning capability under sequential task interaction. The benchmark contains five domain-specific task streams, each consisting of 100 subtasks ordered by difficulty and potential for skill reuse. After completing each subtask, the agent receives feedback and may create or modify skills from its skill repository; the updated repository is then available for subsequent tasks. This setup allows us to examine whether experience from earlier tasks can be consolidated into reusable skills and improve later task performance. Figure~\ref{fig:main} summarizes the benchmark construction, task curation, and evaluation pipeline.

\subsection{Benchmark Construction}

\paragraph{Domain Selection}
\label{subsec:domain_selection}
We select five representative domains in which agents may provide practical value: Healthcare, Law, Mathematics, Finance, and Office. Rather than relying on synthetic environments, these domains span consequential real-world applications and broadly applicable reasoning settings that often require complex, multi-step reasoning or nuanced human evaluation. Their diversity and complexity make them well suited for assessing whether agents can organize accumulated knowledge, refine their problem-solving strategies, and reuse acquired skills over time.

\paragraph{Source Benchmark Selection and Pool Construction}
\label{subsec:source_benchmarks}
To accurately reflect the progressive nature of continual skill learning, we establish a difficulty and capability gradient by constructing our task pool from three distinct tiers of benchmarks (Figure~\ref{fig:bench-composition}). For the foundational layers of our skill chains, we select classic yet challenging datasets (e.g., \textit{OlympiadBench}~\citep{he2024olympiadbenchchallengingbenchmarkpromoting}, \textit{LawBench}~\citep{fei2023lawbenchbenchmarkinglegalknowledge}, \textit{TAT-QA}~\citep{zhu2021tatqaquestionansweringbenchmark}) to test essential knowledge processing skills. We then construct intermediate and advanced stages using open-ended agent datasets (e.g., \textit{GAIA}~\citep{mialon2023gaiabenchmarkgeneralai}, \textit{ClawBench}~\citep{zhang2026clawbenchaiagentscomplete}, \textit{MedAgentsBench}~\citep{tang2025medagentsbenchbenchmarkingthinkingmodels}, \textit{MathCoder}~\citep{wang2024mathcoder}) that demand multi-step reasoning, tool use, and environmental interaction. Finally, for the most challenging stage (e.g., high-level strategic decision-making), we incorporate complex human-evaluated benchmarks like \textit{OneMillionBench (OMBench)}~\citep{yang2026onemillionbenchfarlanguageagents}. Combining these diverse sources ensures our task pool possesses the hierarchical complexity necessary to evaluate an agent's continual self-evolution.

\subsection{Filtering and Ordering via Skill Chains}
\label{subsec:filtering_ordering}

The construction of task sequences in \textsc{ContinualSkillBench} is based on a working assumption: \textit{complex tasks within the same domain often rely on a shared set of foundational skills}. Rather than randomly grouping isolated subtasks from the same domain, we identify the core skills required by each task and organize the tasks according to their potential skill-transfer dependencies. This creates opportunities for skills acquired or refined in one task to be reused across different tasks later in the sequence.

To support progressive learning, we further arrange tasks from relatively simple to more challenging ones. For example, the \textbf{Finance} sequence progresses from basic \textit{financial calculation} to \textit{market analysis} and then to \textit{strategic decision-making}. Similarly, \textbf{Mathematics} moves from \textit{basic algebraic computation} to \textit{mathematical code generation} and finally to \textit{research-level problem solving}. The other domains follow the same general progression: from \textit{medical entity extraction} to \textit{treatment planning} in \textbf{Healthcare}, from \textit{statute retrieval} to \textit{jurisprudential reasoning} in \textbf{Law}, and from \textit{basic email parsing} to \textit{cross-application workflow automation} in \textbf{Office}.

We process the data in three stages:

\begin{enumerate}
    \item \textbf{Skill Labeling and Filtering:}
    We prompt an LLM to identify the skills required by each raw task, filter tasks by their relevance to the target domain skills, and assign an initial difficulty rating.

    \item \textbf{Pairwise Dependency Evaluation and Graph-based Ordering:}
    We sample task pairs from the filtered candidates and ask an LLM to assess the potential direction of skill transfer between them. The resulting judgments are converted into a directed dependency graph. We then order the graph under a difficulty constraint, placing easier tasks first and prioritizing tasks that may support more subsequent tasks.

    \item \textbf{Human Review:}
    Two human annotators reviewed all 500 subtasks for executability and major anomalies in difficulty progression, with no major inconsistency identified. They also audited 400 sampled directional dependency judgments, achieving 89\% exact agreement with the LLM judge.
\end{enumerate}

Further details of the pairwise evaluation and graph-based ordering algorithm are provided in Appendix~\ref{app:dependency_ordering}.

\begin{table*}[!t]
\vspace{-10pt}
\centering
\caption{Performance of Independent (Ind.) and Sequential (Seq.) execution across models and domains. Parentheses report $\Delta=\mathrm{Seq.}-\mathrm{Ind.}$, with \textcolor{ForestGreen}{green} denoting improvement and \textcolor{red}{red} denoting degradation. Raw reward includes all tasks, whereas normalized reward is computed over tasks with valid outputs under both settings.}
\label{tab:main_results}

\begin{subtable}{\textwidth}
\centering
\caption{GPT-4o}
\label{tab:results_gpt4o}
\resizebox{\textwidth}{!}{%
\begin{tabular}{ll ccccc cc}
\toprule
\multirow{2}{*}{\textbf{Domain}} & \multirow{2}{*}{\textbf{Setting}} & \multicolumn{5}{c}{\textbf{Subtask Rewards}} & \multicolumn{2}{c}{\textbf{Reward}} \\
\cmidrule(lr){3-7}\cmidrule(lr){8-9}
& & \textbf{EM} & \textbf{F1} & \textbf{Num.} & \textbf{Prog.} & \textbf{Rubric} & \textbf{Raw} & \textbf{Norm.} \\
\midrule
\cellcolor{LawBG} & Ind. & 0.500 & - & - & 0.300 & 0.111 & 0.286 & 0.325 \\
\multirow{-2}{*}{\cellcolor{LawBG}\textbf{Law}} & Seq. & 0.550~{\color{ForestGreen}(+0.050)} & - & - & 0.300~(0.000) & 0.060~{\color{red}(-0.051)} & 0.280~{\color{red}(-0.006)} & 0.395~{\color{ForestGreen}(+0.070)} \\
\midrule
\cellcolor{FinanceBG} & Ind. & 0.182 & 0.251 & 0.333 & 0.077 & 0.088 & 0.135 & 0.180 \\
\multirow{-2}{*}{\cellcolor{FinanceBG}\textbf{Finance}} & Seq. & 0.182~(0.000) & 0.346~{\color{ForestGreen}(+0.095)} & 0.417~{\color{ForestGreen}(+0.084)} & 0.308~{\color{ForestGreen}(+0.231)} & 0.094~{\color{ForestGreen}(+0.006)} & 0.185~{\color{ForestGreen}(+0.050)} & 0.262~{\color{ForestGreen}(+0.082)} \\
\midrule
\cellcolor{HealthcareBG} & Ind. & 0.107 & - & - & 0.500 & 0.363 & 0.297 & 0.387 \\
\multirow{-2}{*}{\cellcolor{HealthcareBG}\textbf{Healthcare}} & Seq. & 0.429~{\color{ForestGreen}(+0.322)} & - & - & 0.500~(0.000) & 0.417~{\color{ForestGreen}(+0.054)} & 0.424~{\color{ForestGreen}(+0.127)} & 0.413~{\color{ForestGreen}(+0.026)} \\
\midrule
\cellcolor{OfficeBG} & Ind. & 0.000 & - & 0.047 & 0.650 & 0.431 & 0.231 & 0.260 \\
\multirow{-2}{*}{\cellcolor{OfficeBG}\textbf{Office}} & Seq. & 0.000~(0.000) & - & 0.256~{\color{ForestGreen}(+0.209)} & 0.650~(0.000) & 0.443~{\color{ForestGreen}(+0.012)} & 0.308~{\color{ForestGreen}(+0.077)} & 0.380~{\color{ForestGreen}(+0.120)} \\
\midrule
\cellcolor{MathBG} & Ind. & 0.103 & - & 0.208 & 0.000 & 0.040 & 0.100 & 0.110 \\
\multirow{-2}{*}{\cellcolor{MathBG}\textbf{Math}} & Seq. & 0.154~{\color{ForestGreen}(+0.051)} & - & 0.125~{\color{red}(-0.083)} & 0.583~{\color{ForestGreen}(+0.583)} & 0.160~{\color{ForestGreen}(+0.120)} & 0.200~{\color{ForestGreen}(+0.100)} & 0.198~{\color{ForestGreen}(+0.088)} \\
\bottomrule
\end{tabular}%
}
\end{subtable}

\vspace{0.8em}

\begin{subtable}{\textwidth}
\centering
\caption{GPT-5.3-Codex}
\label{tab:results_gpt53}
\resizebox{\textwidth}{!}{%
\begin{tabular}{ll ccccc cc}
\toprule
\multirow{2}{*}{\textbf{Domain}} & \multirow{2}{*}{\textbf{Setting}} & \multicolumn{5}{c}{\textbf{Subtask Rewards}} & \multicolumn{2}{c}{\textbf{Reward}} \\
\cmidrule(lr){3-7}\cmidrule(lr){8-9}
& & \textbf{EM} & \textbf{F1} & \textbf{Num.} & \textbf{Prog.} & \textbf{Rubric} & \textbf{Raw} & \textbf{Norm.} \\
\midrule
\cellcolor{LawBG} & Ind. & 0.875 & - & - & 0.800 & 0.309 & 0.585 & 0.585 \\
\multirow{-2}{*}{\cellcolor{LawBG}\textbf{Law}} & Seq. & 0.975~{\color{ForestGreen}(+0.100)} & - & - & 0.800~(0.000) & 0.319~{\color{ForestGreen}(+0.010)} & 0.629~{\color{ForestGreen}(+0.044)} & 0.629~{\color{ForestGreen}(+0.044)} \\
\midrule
\cellcolor{FinanceBG} & Ind. & 0.364 & 0.309 & 0.417 & 0.800 & 0.404 & 0.435 & 0.432 \\
\multirow{-2}{*}{\cellcolor{FinanceBG}\textbf{Finance}} & Seq. & 0.455~{\color{ForestGreen}(+0.091)} & 0.506~{\color{ForestGreen}(+0.197)} & 0.833~{\color{ForestGreen}(+0.416)} & 0.900~{\color{ForestGreen}(+0.100)} & 0.442~{\color{ForestGreen}(+0.038)} & 0.495~{\color{ForestGreen}(+0.060)} & 0.556~{\color{ForestGreen}(+0.124)} \\
\midrule
\cellcolor{HealthcareBG} & Ind. & 0.321 & - & - & 0.500 & 0.396 & 0.379 & 0.380 \\
\multirow{-2}{*}{\cellcolor{HealthcareBG}\textbf{Healthcare}} & Seq. & 0.857~{\color{ForestGreen}(+0.536)} & - & - & 0.500~(0.000) & 0.521~{\color{ForestGreen}(+0.125)} & 0.614~{\color{ForestGreen}\textbf{(+0.235)}} & 0.620~{\color{ForestGreen}\textbf{(+0.240)}} \\
\midrule
\cellcolor{OfficeBG} & Ind. & 1.000 & - & 0.860 & 1.000 & 0.367 & 0.718 & 0.731 \\
\multirow{-2}{*}{\cellcolor{OfficeBG}\textbf{Office}} & Seq. & 0.500~{\color{red}(-0.500)} & - & 0.881~{\color{ForestGreen}(+0.021)} & 1.000~(0.000) & 0.408~{\color{ForestGreen}(+0.041)} & 0.723~{\color{ForestGreen}(+0.005)} & 0.738~{\color{ForestGreen}(+0.007)} \\
\midrule
\cellcolor{MathBG} & Ind. & 0.538 & - & 0.708 & 0.667 & 0.840 & 0.670 & 0.670 \\
\multirow{-2}{*}{\cellcolor{MathBG}\textbf{Math}} & Seq. & 0.590~{\color{ForestGreen}(+0.052)} & - & 0.875~{\color{ForestGreen}(+0.167)} & 0.750~{\color{ForestGreen}(+0.083)} & 0.860~{\color{ForestGreen}(+0.020)} & 0.745~{\color{ForestGreen}(+0.075)} & 0.745~{\color{ForestGreen}(+0.075)} \\
\bottomrule
\end{tabular}%
}
\end{subtable}

\vspace{0.8em}

\begin{subtable}{\textwidth}
\centering
\caption{Opus 4.7}
\label{tab:results_opus47}
\resizebox{\textwidth}{!}{%
\begin{tabular}{ll ccccc cc}
\toprule
\multirow{2}{*}{\textbf{Domain}} & \multirow{2}{*}{\textbf{Setting}} & \multicolumn{5}{c}{\textbf{Subtask Rewards}} & \multicolumn{2}{c}{\textbf{Reward}} \\
\cmidrule(lr){3-7}\cmidrule(lr){8-9}
& & \textbf{EM} & \textbf{F1} & \textbf{Num.} & \textbf{Prog.} & \textbf{Rubric} & \textbf{Raw} & \textbf{Norm.} \\
\midrule
\cellcolor{LawBG} & Ind. & 0.850 & - & - & 0.800 & 0.269 & 0.556 & 0.573 \\
\multirow{-2}{*}{\cellcolor{LawBG}\textbf{Law}} & Seq. & 0.950~{\color{ForestGreen}(+0.100)} & - & - & 0.800~(0.000) & 0.307~{\color{ForestGreen}(+0.038)} & 0.595~{\color{ForestGreen}(+0.039)} & 0.633~{\color{ForestGreen}(+0.060)} \\
\midrule
\cellcolor{FinanceBG} & Ind. & 0.364 & 0.381 & 0.833 & 0.769 & 0.259 & 0.414 & 0.438 \\
\multirow{-2}{*}{\cellcolor{FinanceBG}\textbf{Finance}} & Seq. & 0.364~(0.000) & 0.476~{\color{ForestGreen}(+0.095)} & 0.750~{\color{red}(-0.083)} & 0.769~(0.000) & 0.302~{\color{ForestGreen}(+0.043)} & 0.442~{\color{ForestGreen}(+0.028)} & 0.460~{\color{ForestGreen}(+0.022)} \\
\midrule
\cellcolor{HealthcareBG} & Ind. & 0.750 & - & - & 0.500 & 0.282 & 0.422 & 0.481 \\
\multirow{-2}{*}{\cellcolor{HealthcareBG}\textbf{Healthcare}} & Seq. & 0.929~{\color{ForestGreen}(+0.179)} & - & - & 0.500~(0.000) & 0.516~{\color{ForestGreen}(+0.234)} & 0.631~{\color{ForestGreen}(+0.209)} & 0.662~{\color{ForestGreen}(+0.181)} \\
\midrule
\cellcolor{OfficeBG} & Ind. & 1.000 & - & 0.884 & 1.000 & 0.305 & 0.707 & 0.707 \\
\multirow{-2}{*}{\cellcolor{OfficeBG}\textbf{Office}} & Seq. & 1.000~(0.000) & - & 0.884~(0.000) & 1.000~(0.000) & 0.405~{\color{ForestGreen}(+0.100)} & 0.742~{\color{ForestGreen}(+0.035)} & 0.742~{\color{ForestGreen}(+0.035)} \\
\midrule
\cellcolor{MathBG} & Ind. & 0.641 & - & 0.917 & 0.750 & 0.400 & 0.660 & 0.660 \\
\multirow{-2}{*}{\cellcolor{MathBG}\textbf{Math}} & Seq. & 0.744~{\color{ForestGreen}(+0.103)} & - & 0.917~(0.000) & 0.750~(0.000) & 0.208~{\color{red}(-0.192)} & 0.652~{\color{red}(-0.008)} & 0.652~{\color{red}(-0.008)} \\
\bottomrule
\end{tabular}
}
\end{subtable}
\end{table*}

\begin{table*}[h]
\centering
\caption{Performance under Independent (Ind.), pure ICL, and
skill-maintaining Sequential (Seq.) execution. Parentheses report
$\Delta=\mathrm{Seq.}-\mathrm{ICL}$, with
\textcolor{ForestGreen}{green} denoting improvement and
\textcolor{red}{red} denoting degradation.}
\label{tab:icl_seq_results}

\begin{subtable}{\textwidth}
\centering
\caption{GPT-5.3-Codex}
\label{tab:icl_seq_codex}
\resizebox{\textwidth}{!}{%
\begin{tabular}{ll ccccc cc}
\toprule
\multirow{2}{*}{\textbf{Domain}}
& \multirow{2}{*}{\textbf{Setting}}
& \multicolumn{5}{c}{\textbf{Subtask Rewards}}
& \multicolumn{2}{c}{\textbf{Reward}} \\
\cmidrule(lr){3-7}\cmidrule(lr){8-9}
& & \textbf{EM} & \textbf{F1} & \textbf{Num.}
& \textbf{Prog.} & \textbf{Rubric}
& \textbf{Raw} & \textbf{Norm.} \\
\midrule
\cellcolor{LawBG} & Ind.
& 0.875 & -- & -- & 0.800 & 0.309 & 0.585 & 0.585 \\
\cellcolor{LawBG} & ICL
& 0.900 & -- & -- & 0.900 & 0.330 & 0.615 & 0.617 \\
\multirow{-3}{*}{\cellcolor{LawBG}\textbf{Law}}
& Seq.
& 0.975~{\color{ForestGreen}(+0.075)}
& --
& --
& 0.800~{\color{red}(-0.100)}
& 0.319~{\color{red}(-0.011)}
& 0.629~{\color{ForestGreen}(+0.014)}
& 0.629~{\color{ForestGreen}(+0.012)} \\
\midrule
\cellcolor{FinanceBG} & Ind.
& 0.364 & 0.309 & 0.417 & 0.800 & 0.404 & 0.435 & 0.432 \\
\cellcolor{FinanceBG} & ICL
& 0.364 & 0.460 & 0.833 & 0.900 & 0.444 & 0.482 & 0.542 \\
\multirow{-3}{*}{\cellcolor{FinanceBG}\textbf{Finance}}
& Seq.
& 0.455~{\color{ForestGreen}(+0.091)}
& 0.506~{\color{ForestGreen}(+0.046)}
& 0.833~(0.000)
& 0.900~(0.000)
& 0.442~{\color{red}(-0.002)}
& 0.495~{\color{ForestGreen}(+0.013)}
& 0.556~{\color{ForestGreen}(+0.014)} \\
\midrule
\cellcolor{HealthcareBG} & Ind.
& 0.321 & -- & -- & 0.500 & 0.396 & 0.379 & 0.380 \\
\cellcolor{HealthcareBG} & ICL
& 0.857 & -- & -- & 0.250 & 0.596 & 0.655 & 0.655 \\
\multirow{-3}{*}{\cellcolor{HealthcareBG}\textbf{Healthcare}}
& Seq.
& 0.857~(0.000)
& --
& --
& 0.500~{\color{ForestGreen}(+0.250)}
& 0.521~{\color{red}(-0.075)}
& 0.614~{\color{red}(-0.041)}
& 0.620~{\color{red}(-0.035)} \\
\bottomrule
\end{tabular}%
}
\end{subtable}

\vspace{0.8em}

\begin{subtable}{\textwidth}
\centering
\caption{Opus 4.7}
\label{tab:icl_seq_opus}
\resizebox{\textwidth}{!}{%
\begin{tabular}{ll ccccc cc}
\toprule
\multirow{2}{*}{\textbf{Domain}}
& \multirow{2}{*}{\textbf{Setting}}
& \multicolumn{5}{c}{\textbf{Subtask Rewards}}
& \multicolumn{2}{c}{\textbf{Reward}} \\
\cmidrule(lr){3-7}\cmidrule(lr){8-9}
& & \textbf{EM} & \textbf{F1} & \textbf{Num.}
& \textbf{Prog.} & \textbf{Rubric}
& \textbf{Raw} & \textbf{Norm.} \\
\midrule
\cellcolor{LawBG} & Ind.
& 0.850 & -- & -- & 0.800 & 0.272 & 0.556 & 0.556 \\
\cellcolor{LawBG} & ICL
& 0.925 & -- & -- & 0.700 & 0.366 & 0.608 & 0.633 \\
\multirow{-3}{*}{\cellcolor{LawBG}\textbf{Law}}
& Seq.
& 0.950~{\color{ForestGreen}(+0.025)}
& --
& --
& 0.800~{\color{ForestGreen}(+0.100)}
& 0.270~{\color{red}(-0.096)}
& 0.595~{\color{red}(-0.013)}
& 0.595~{\color{red}(-0.038)} \\
\midrule
\cellcolor{FinanceBG} & Ind.
& 0.364 & 0.381 & 0.833 & 0.769 & 0.259 & 0.414 & 0.438 \\
\cellcolor{FinanceBG} & ICL
& 0.273 & 0.505 & 0.833 & 0.700 & 0.218 & 0.398 & 0.403 \\
\multirow{-3}{*}{\cellcolor{FinanceBG}\textbf{Finance}}
& Seq.
& 0.364~{\color{ForestGreen}(+0.091)}
& 0.476~{\color{red}(-0.029)}
& 0.750~{\color{red}(-0.083)}
& 0.769~{\color{ForestGreen}(+0.069)}
& 0.302~{\color{ForestGreen}(+0.084)}
& 0.442~{\color{ForestGreen}(+0.044)}
& 0.460~{\color{ForestGreen}(+0.057)} \\
\midrule
\cellcolor{HealthcareBG} & Ind.
& 0.750 & -- & -- & 0.500 & 0.282 & 0.422 & 0.422 \\
\cellcolor{HealthcareBG} & ICL
& 0.893 & -- & -- & 0.500 & 0.571 & 0.658 & 0.658 \\
\multirow{-3}{*}{\cellcolor{HealthcareBG}\textbf{Healthcare}}
& Seq.
& 0.929~{\color{ForestGreen}(+0.036)}
& --
& --
& 0.500~(0.000)
& 0.516~{\color{red}(-0.055)}
& 0.631~{\color{red}(-0.027)}
& 0.631~{\color{red}(-0.027)} \\
\bottomrule
\end{tabular}%
}
\end{subtable}

\end{table*}

\subsection{Structural Validation of Skill Continuity}
\label{subsec:skill_continuity}

To verify that the resulting sequences provide meaningful opportunities for cross-task transfer, we annotate each task's core skills independently of its sequence position. A provenance-blinded human study on 50 tasks supports the reliability of these annotations: averaged across two independent annotators, 92.6\% of the proposed core skills were accepted, and both annotators found at least mostly complete skill coverage for 49 of 50 tasks.

Using embedding-based semantic matching at a cosine threshold of 0.85, 69.5\% of eligible tasks reuse at least one previously encountered core skill, while 35.5\% of their core-skill requirements are covered by the preceding history, macro-averaged across domains. The curated sequences also achieve higher local coverage than random task permutations in all five domains and for history windows of 1, 5, and 10 tasks.

We further repeat the analysis using Gemma-3-12B as an independent functional-equivalence judge. For a five-task window, the curated sequences achieve 25.7\% coverage versus 20.4\% under random permutations (\(+5.3\) percentage points), with all five domains remaining significant after Holm correction. This conclusion holds despite the low overlap between the embedding- and Gemma-based equivalence graphs (Jaccard \(=0.180\)), indicating that the observed local skill continuity is not an artifact of a particular matching mechanism but a robust structural property of the curated task sequences. Full protocols, robustness analyses, and domain-level results are provided in Appendix~\ref{app:skill_continuity}.

\subsection{Evaluation Metrics}
\label{subsec:evaluation_metrics}
Because the benchmark includes tasks with different output formats and evaluation requirements, we use four types of evaluators:

\begin{itemize}
    \item \textbf{Exact Match and F1}: Exact match is used for tasks with a short, deterministic answer. For longer responses, token-level F1 measures partial overlap with the reference answer.

    \item \textbf{Numeric}: Numerical answers are extracted and compared with the reference value under a task-specific tolerance, typically $\epsilon \leq 10^{-4}$.

    \item \textbf{Rubric Judge}: Open-ended responses are evaluated by an LLM judge using predefined rubric criteria. Task reward is the weighted aggregate of the criterion-level scores.

    \item \textbf{Programmatic}: Tasks involving tool use or artifact creation are evaluated with executable tests that inspect the resulting files, outputs, or environment states. Where available, we adapt the official evaluators from the source benchmarks, including \textit{GAIA} and \textit{ClawBench}.
\end{itemize}


\subsection{Underlying Framework}
\label{subsec:underlying_framework}

We build \textsc{ContinualSkillBench} on the Harbor infrastructure
~\citep{Harbor_Framework,merrill2026terminalbenchbenchmarkingagentshard}
and extend it with two sequential agent harnesses based on Codex CLI
~\citep{openai2025codexcli} and Claude Code
~\citep{anthropic2025claudecode}. Each subtask follows a three-turn
protocol: the agent first receives the task and its current skill
repository, then executes the task, and finally reflects on evaluator
feedback. During reflection, it may create or revise skills using the
provided \textbf{Create Skill} and \textbf{Modify Skill} meta-skills.
Any updates become available from the next subtask onward. The complete
interaction protocol and prompts are provided in
Appendix~\ref{app:prompts}.

\section{Experiments}

\subsection{Environmental Setup}

\paragraph{Evaluated Models.}
To assess the state-of-the-art agentic capabilities under our sequential setting, we evaluate three representative foundation models on \textsc{ContinualSkillBench}: GPT-4o ~\citep{openai2024gpt4ocard}, GPT-5.3-Codex ~\citep{openai20265.3codex}, and Claude 4.7 Opus ~\citep{anthropic2026claudeopus47}. All models use the agent harnesses described in Section~\ref{subsec:underlying_framework}, with identical task orders and evaluation procedures within each domain.

\paragraph{Evaluation Settings.}
We consider the following two primary settings:

\begin{itemize}
    \item \textbf{Independent Execution (Ind., Baseline):} 
    The agent solves each task independently. Its interaction history and skill repository are reset before every task.
    \item \textbf{Sequential Execution (Seq.):} The agent processes the 100 subtasks sequentially, actively retaining and updating its skill repository throughout the entire trajectory.
\end{itemize}

Sequential--Independent differences capture the combined effects of retained context, feedback, and explicit skill maintenance. We further isolate the latter using the pure-ICL baseline introduced in Section~\ref{sec:icl_ablation}.

\paragraph{Aggregate Metrics.}






We report two complementary metrics. Raw reward
\(R_{\mathrm{raw}}\) averages scores over all 100 tasks and therefore
captures both solution quality and execution reliability. Normalized
reward \(R_{\mathrm{norm}}\) is computed on the common subset for
which both settings produce valid outputs, providing a diagnostic of
solution quality conditional on successful execution. Because this
intersection may exclude setting-specific failures, we interpret
normalized reward jointly with raw reward.

For \(m\in\{\mathrm{raw},\mathrm{norm}\}\), the absolute
Sequential--Independent gain is
\vspace{-4pt}
\begin{equation}
    \Delta_m = R_m^{\mathrm{seq}} - R_m^{\mathrm{ind}}.
\end{equation}

Because the common valid-output subset is defined over all settings in each comparison, normalized rewards may differ slightly across experiments involving different sets of settings.

\subsection{Main Results}
\label{subsec:main_results}

Table~\ref{tab:main_results} presents the main evaluation results. We organize the analysis around four observations.

\begin{figure*}[t]
    \centering
    \vspace{-5pt}
    \begin{subfigure}[t]{0.49\linewidth}
        \centering
        \includegraphics[width=\linewidth]{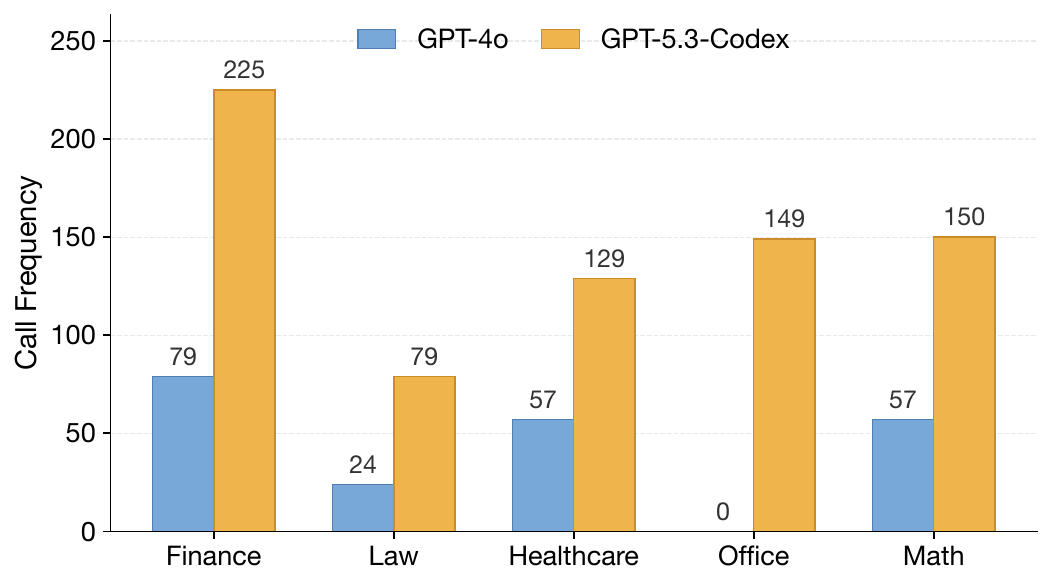}
        \caption{Skill call frequency}
        \label{fig:skill-call-frequency}
    \end{subfigure}
    \hfill
    \begin{subfigure}[t]{0.49\linewidth}
        \centering
        \includegraphics[width=\linewidth]{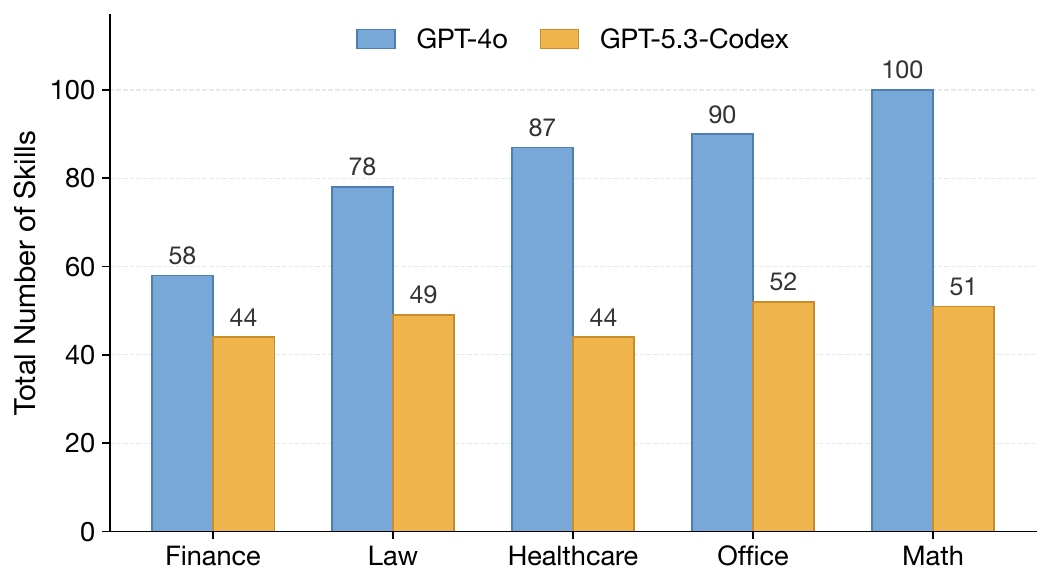}
        \caption{Final skill pool size}
        \label{fig:skill-pool-size}
    \end{subfigure}
    \caption{Skill-library behavior of GPT-4o and GPT-5.3-Codex across five domains. }
    \vspace{-8pt}
    \label{skill-results}
\end{figure*}

\paragraph{Sequential execution generally improves task performance.}
Sequential execution increases raw reward in 13 of the 15 model--domain combinations and normalized reward in 14 of 15. Macro-averaged across all combinations, the absolute improvements are $+0.071$ in raw reward and $+0.078$ in normalized reward. Relative to the corresponding aggregate Independent baselines, these changes represent improvements of 16.2\% and 16.9\%, respectively. Thus, retaining experience across tasks is generally beneficial, although
the improvement is not universal.

\paragraph{The magnitude of improvement varies across models.} GPT-5.3-Codex obtains the largest average normalized improvement ($+0.098$), followed by GPT-4o ($+0.077$) and Opus 4.7 ($+0.058$). This ordering does not directly follow Independent performance: although Opus 4.7 has the strongest average Independent baseline, it does not obtain the largest sequential gain. Conversely, GPT-4o achieves substantial relative improvements from a lower baseline. These results suggest that the ability to benefit from prior experience is model-dependent but is not determined solely by baseline task-solving capability.

\paragraph{Sequential benefits are strongly domain-dependent.}
Healthcare exhibits the largest average normalized improvement across models ($+0.149$), driven primarily by gains from GPT-5.3-Codex and Opus 4.7. The average gains in Finance, Law, Office, and Mathematics are $+0.076$, $+0.058$, $+0.054$, and $+0.052$, respectively. The only decrease in normalized reward occurs for Opus 4.7 on Mathematics ($-0.008$). 

\paragraph{Improvements also depend on the evaluator type.}
Structured tasks often benefit from sequential execution. For example, GPT-5.3-Codex improves by $+0.416$ on Numeric tasks and $+0.091$ on Exact Match tasks in Finance, compared with $+0.038$ on Rubric-judged tasks. However, this pattern is not universal: Opus 4.7 obtains a substantial $+0.234$ Rubric improvement in Healthcare, while its Mathematics Rubric score decreases by $-0.192$. These differences motivate a closer examination of whether the gains arise from explicit skill maintenance or from broader adaptation to previous context and feedback.

\subsection{Explicit Skill Maintenance vs. Pure In-Context Learning}
\label{sec:icl_ablation}

Sequential--Independent comparison does not isolate the contribution of the explicit skill repository, since Sequential execution also retains prior context and feedback. We therefore introduce a pure in-context learning (ICL) condition that follows the same sequence and \textbf{receives the same context and evaluator feedback, but cannot create or modify the skill repository during the sequence}. Accordingly, Independent--ICL captures general adaptation through retained interactions and feedback, whereas ICL--Sequential isolates the contribution of explicit skill maintenance beyond these factors. We evaluate this condition with GPT-5.3-Codex and Claude 4.7 Opus on Law, Finance, and Healthcare.

Macro-averaged across the six model--domain combinations, the normalized rewards of Independent, ICL, and skill-maintaining Sequential execution are 0.469, 0.585, and 0.582, respectively. The near-identical aggregate performance of ICL and Sequential, together with variation across individual models and domains, shows that explicit skill maintenance provides no consistent aggregate advantage over retaining context and feedback alone.

The two conditions nevertheless exhibit consistent task-type differences across models. Sequential improves Exact Match performance in five of the six comparisons and ties ICL in the remaining one; it also matches or outperforms ICL on Programmatic tasks in five of six comparisons. Conversely, ICL obtains higher Rubric scores in five of six comparisons. These results suggest that much of the Sequential--Independent gain comes from retained context and feedback. Explicit skills can stabilize reusable procedures under rigid output or execution requirements, but may over-specialize to earlier solution patterns on open-ended tasks.

\subsection{Skill Library Dynamics}
\label{subsec:skill_dynamics}

Performance gains alone do not reveal how agents maintain their skill repositories. We therefore examine two complementary behavioral statistics: the frequency with which generated skills are invoked in later tasks and the size of the resulting skill repository. Figure~\ref{skill-results} reports these statistics for GPT-4o and GPT-5.3-Codex across the five domains.

\paragraph{Skill fragmentation hinders long-term adaptation in weaker
models.}
Figure~\ref{skill-results} reveals a clear contrast in how the two models maintain their skill repositories. GPT-5.3-Codex constructs a
relatively compact library while invoking its skills more frequently in subsequent tasks. This pattern suggests that it can consolidate multiple task experiences into reusable procedures and retrieve them when related tasks reappear. GPT-4o exhibits the opposite behavior: it accumulates a larger skill pool but invokes the resulting skills less frequently.

This combination of rapid library growth and limited reuse indicates that GPT-4o tends to preserve fragmented, task-specific skills rather than merging related experience into more general procedures. As the trajectory progresses, these narrowly scoped skills enlarge the repository without providing proportional downstream utility, increasing the burden of selecting and maintaining useful skills. Additional analyses of generated-skill
validity and content quality are provided in
Appendix~\ref{app:generated skills}.

\section{Conclusion}


We introduced \textsc{ContinualSkillBench} to evaluate continual skill learning across long task sequences in five domains. Sequential execution generally improves performance, but the gains vary across models and domains. ICL performs comparably to explicit skill maintenance on average, indicating that much of the improvement comes from retained context and feedback. Explicit skills better support rigid outputs and fixed procedures, but may constrain performance on open-ended tasks. Weaker models build larger, more fragmented repositories yet rarely reuse their generated skills, suggesting limited skill generalizability. Overall, current agents benefit from sequential interaction but still struggle to consolidate experience into transferable skills.




\section*{Limitations}

Our study has several limitations. First, while \textsc{ContinualSkillBench} covers five core domains, the tasks within each domain are still curated from a fixed set of sources. Real deployments are more diverse and less controlled: agents may see rare edge cases, shifted data distributions, and instructions that differ substantially from the benchmark format. These long-tail settings are not fully covered by our current evaluation.

Second, due to the high time and API cost of sequential agent evaluation, we evaluate only a limited set of representative models and harnesses. We do not exhaustively cover additional Claude, GPT, or Gemini variants, nor do we adapt the benchmark to other agent environments such as Cursor or Google CLI. Since different models and agent frameworks may handle memory, skill retrieval, and execution control differently, future work should broaden both model and infrastructure coverage.


\bibliography{custom}

\appendix
\section{Detailed Task Sources}
\label{sec:appendix_task_sources}

In this section, we provide a comprehensive overview of the primary data sources integrated into \textsc{ContinualSkillBench}. To construct a robust evaluation suite, we combine cross-domain foundational benchmarks with specialized, domain-specific datasets. Table~\ref{tab:task_sources_summary} summarizes the datasets utilized and their respective target capabilities.

\begin{table*}[htbp]
\centering
\small
\setlength{\tabcolsep}{6pt}
\begin{tabular}{llp{9.5cm}}
\toprule
\textbf{Domain} & \textbf{Source Dataset} & \textbf{Core Capability / Subtask Focus} \\
\midrule
\multirow{3}{*}{\textbf{Cross-Domain}} 
 & ClawBench & Sequential workflows, tool manipulation, and multi-step execution \\
 & OneMillionBench & Large-scale information extraction, text processing, and document comprehension \\
 & GAIA & Complex multimodal assistant tasks and real-world reasoning \\
\midrule
\multirow{3}{*}{\textbf{Finance}} 
 & InvestorBench & Investment evaluation, quantitative analysis, and portfolio optimization \\
 & TAT-QA & Hybrid tabular-textual financial reasoning and numerical computation \\
 & FinBen & Comprehensive core financial capability and knowledge evaluation \\
\midrule
\multirow{3}{*}{\textbf{Healthcare}} 
 & MedAgentsBench & Multi-agent medical consultation and collective diagnostic reasoning \\
 & HealthBench & Public health question answering and consumer medical understanding \\
 & PhysicianBench & Professional clinical diagnostic reasoning and differential analysis \\
\midrule
\multirow{4}{*}{\textbf{Law}} 
 & LegalBench & Collaborative legal reasoning, contract classification, and rule application \\
 & LawBench & Legal knowledge alignment, statutory understanding, and case analysis \\
 & harvey-labs & Professional-grade legal analysis, citation matching, and draft auditing \\
 & Lexam & Bar examination comprehension and statutory interpretation \\
\midrule
\multirow{8}{*}{\textbf{Math}} 
 & OlympiadBench & High-level competition mathematics and advanced problem solving \\
 & amc12 & American Mathematics Competitions (Grade 12) reasoning tracks \\
 & aime & American Invitational Mathematics Examination challenge cases \\
 & NuminaMath & Diverse chain-of-thought mathematical reasoning and logic tracing \\
 & OmniMath & Comprehensive multi-discipline mathematical suite covering diverse topics \\
 & Putnam\_Axiom & Research-level axiomatic mathematics and rigorous Putnam-level proofs \\
 & MathCoder & Mathematics-oriented code generation, execution, and self-debugging \\
 & RealMath & Grounded real-world numerical problem solving and applied math scaling \\
\midrule
\multirow{2}{*}{\textbf{Office}} 
 & OfficeQA & Paragraph-level question answering over complex enterprise documents \\
 & WritingBench & Professional document drafting, stylistic refinement, and formatting \\
\bottomrule
\end{tabular}
\caption{Summary of source datasets and their corresponding target capabilities in \textsc{ContinualSkillBench}, categorized into cross-domain foundational benchmarks and domain-specific suites.}
\label{tab:task_sources_summary}
\end{table*}

\subsection{Dataset Descriptions}

\paragraph{Cross-Domain Benchmarks}
To establish a foundational baseline for general assistant capabilities, we integrate several comprehensive benchmarks across multiple domains (Finance, Healthcare, Law, and Office). Specifically, \textbf{ClawBench} ~\citep{zhang2026clawbenchaiagentscomplete} is utilized to evaluate multi-step workflow execution and cross-application tool manipulation. \textbf{OneMillionBench} \citep{yang2026onemillionbenchfarlanguageagents} provides high-throughput tasks for large-scale text processing and entity extraction. Furthermore, \textbf{GAIA} \citep{mialon2023gaiabenchmarkgeneralai} is incorporated to challenge models with complex, real-world multimodal reasoning and long-horizon assistant workflows.

\paragraph{Finance}
Beyond general text processing, the financial tasks in \textsc{ContinualSkillBench} are designed to evaluate high-precision numerical accuracy and strategic decision-making. We integrate data from \textbf{TAT-QA} \citep{zhu2021tatqaquestionansweringbenchmark} to challenge the model's hybrid table-text question answering, while \textbf{InvestorBench} \citep{li2024investorbenchbenchmarkfinancialdecisionmaking} and \textbf{FinBen} \citep{xie2024finbenholisticfinancialbenchmark} provide professional-grade foundations for investment analysis, portfolio optimization, and core financial knowledge.

\paragraph{Healthcare}
Evaluating clinical assistants requires a rigorous combination of specialized medical knowledge and safe diagnostic reasoning. We adapt professional clinical reasoning scenarios from \textbf{PhysicianBench} \citep{liu2026physicianbenchevaluatingllmagents}, multi-agent consultation setups from \textbf{MedAgentsBench} \citep{tang2025medagentsbenchbenchmarkingthinkingmodels}, and public health scenarios from \textbf{HealthBench} \citep{arora2025healthbenchevaluatinglargelanguage}. These domain-specific tasks ensure robust evaluation of clinical safety and collective diagnostic accuracy.

\paragraph{Law}
Legal evaluation within our benchmark hinges on absolute textual precision, statutory interpretation, and rigorous clause matching. We draw heavily from \textbf{LegalBench} \citep{guha2023legalbenchcollaborativelybuiltbenchmark} for structured legal reasoning, and incorporate localized legal knowledge using \textbf{LawBench} \citep{fei2023lawbenchbenchmarkinglegalknowledge}. Professional legal intelligence data from \textbf{harvey-labs} \citep{grupen2026harveylab} and legal examination tracks from \textbf{Lexam} \citep{fan2026lexambenchmarkinglegalreasoning} are embedded to test advanced analysis and citation matching.

\paragraph{Math}
Mathematical logic is evaluated up to competition and research-level intensities. We source highly sophisticated problems from \textbf{OlympiadBench} \citep{gao2024omnimathuniversalolympiadlevel}, \textbf{amc12} \citep{edev2000amc12}, \textbf{aime} \citep{emergentmind2026aime}, and the axiomatic proof tracks of \textbf{Putnam\_Axiom} \citep{gulati2025putnamaxiomfunctionalstaticbenchmark}. Broader multi-discipline mathematical reasoning paths are enriched via \textbf{NuminaMath} \citep{numina_math_datasets} and \textbf{OmniMath} \citep{gao2024omnimathuniversalolympiadlevel}. Additionally, code-driven mathematical solving and execution-based validation environments are established through the integration of \textbf{MathCoder} \citep{wang2024mathcoder, zhou2024solving} and \textbf{RealMath} \citep{zhang2025realmathcontinuousbenchmarkevaluating}.

\paragraph{Office}
Office tasks simulate daily desktop automation and specialized enterprise document processing. We utilize \textbf{OfficeQA} \citep{opsahlong2026officeqaproenterprisebenchmark} for paragraph-level question answering and information retrieval over complex documents, combined with \textbf{WritingBench} \citep{wu2025writingbenchcomprehensivebenchmarkgenerative} to evaluate professional content generation, stylistic refinement, and formatting capabilities.

\section{Pairwise Dependency Evaluation and Graph-based Ordering}
\label{app:dependency_ordering}

\subsection{Pairwise Dependency Evaluation}

For each domain, we sample 200 unordered pairs from the 100 filtered tasks. Each pair is evaluated in both directions, producing 400 directional judgments per domain. Given a pair of tasks $(A,B)$, the LLM (specifically GPT-5.4) determines whether completing Task~$A$ could provide skills useful for Task~$B$, and vice versa. Each direction is assigned one of three labels: \texttt{YES}, \texttt{PARTIAL}, or \texttt{NO}.

A \texttt{YES} judgment indicates a clear potential transfer relationship and is used to construct the dependency graph. \texttt{PARTIAL} and \texttt{NO} judgments are retained for analysis but are not used as graph edges.



The prompt we use is as follows:

\begin{tcolorbox}[
    colback=gray!3,
    colframe=gray!50,
    sharp corners,
    coltitle=black,
    fonttitle=\bfseries,
    boxrule=0.5pt,
    left=6pt,
    right=6pt,
    top=6pt,
    bottom=6pt,
    fontupper=\ttfamily\small,
    title={Skill Transfer Judge Prompt}
]
You are evaluating whether two tasks share transferable skills that an AI agent could learn.

\vspace{\baselineskip}
\textbf{Input tasks}

Task A: \{task\_a['question']\}\\
Task B: \{task\_b['question']\}

\vspace{\baselineskip}
\textbf{Instruction}

Analyze the skill transfer relationship between these two tasks. Consider:

1. Prerequisite sub-skills: Does Task A require sub-skills (e.g., reading tables, ratio calculation, parsing text) that also appear in Task B? Or vice versa?\\
2. Shared skill practice: Do both tasks exercise the same underlying skill, meaning doing either one gives useful practice for the other?\\
3. Transfer direction: Is the transfer primarily A->B (doing A first helps B), B->A (doing B first helps A), or bidirectional?

\vspace{\baselineskip}
\textbf{Required output}

Respond in this exact JSON format:\\
\{
  "a\_helps\_b": "YES" | "PARTIAL" | "NO",\\
  "b\_helps\_a": "YES" | "PARTIAL" | "NO",\\
  "shared\_skills": ["skill1", "skill2"],\\
  "reason": "one sentence explaining the main skill relationship"\\
\}
\end{tcolorbox}

\begin{table}[H]
\centering
\caption{Pairwise dependency judgments across the five domains. Each
sampled pair is evaluated in both directions, producing 400 directional
judgments per domain.}
\label{tab:dependency_evaluation}
\small
\setlength{\tabcolsep}{3.5pt}
\renewcommand{\arraystretch}{0.88}
\begin{tabular}{lccccc}
\toprule
\textbf{Domain}
& \textbf{Tasks}
& \textbf{Pairs}
& \textbf{Yes}
& \textbf{Partial}
& \textbf{No} \\
\midrule
Finance    & 100 & 200 & 112 & 96  & 192 \\
Law        & 100 & 200 & 94  & 117 & 189 \\
Office     & 100 & 200 & 103 & 108 & 189 \\
Healthcare & 100 & 200 & 121 & 102 & 177 \\
Math       & 100 & 200 & 76  & 134 & 190 \\
\bottomrule
\end{tabular}
\end{table}

\begin{table}[H]
\centering
\caption{Human validation of sampled directional dependency judgments. Label counts report the GPT-5.4 judgments within the audited sample.}
\label{tab:dependency_human_validation}
\small
\setlength{\tabcolsep}{5pt}
\begin{tabular}{lrrrr}
\toprule
\textbf{Domain}
& \textbf{Yes}
& \textbf{Partial}
& \textbf{No}
& \textbf{Exact Agreement} \\
\midrule
Finance & 99  & 84  & 17 & 190/200 (95.0\%) \\
Law     & 76  & 95  & 29 & 166/200 (83.0\%) \\
\midrule
Overall & 175 & 179 & 46 & 356/400 (89.0\%) \\
\bottomrule
\end{tabular}
\end{table}

\subsection{Graph-based Ordering}

We construct a directed graph $G=(V,E)$ from the pairwise judgments. Each node in $V$ represents a task, and an edge $A\rightarrow B$ is added when the LLM assigns \texttt{YES} to the corresponding transfer direction. An edge therefore indicates that Task~$A$ is expected to provide skills useful for Task~$B$.

The final task order is generated under a curriculum constraint. Tasks are first grouped by difficulty level and processed from easier to more challenging levels. For Mathematics, we use the predefined skill layers as the curriculum levels. Edges that point from a higher curriculum level to a lower one are removed so that they do not conflict with the intended progression.

Within each curriculum level, we apply a greedy variant of Kahn's topological sorting algorithm. At each step, we identify tasks with no remaining incoming edges and select the one with the largest outgoing degree. This prioritizes tasks that are expected to provide useful skills for more downstream tasks. If the graph contains a cycle and no zero-in-degree task is available, we break the cycle by selecting the remaining task with the largest outgoing degree. Finally, the ordered curriculum levels are concatenated to form the complete 100-task sequence.

\subsection{Human Validation Protocol}

Two human annotators reviewed all 500 subtasks for task executability. For each subtask, they checked whether the instruction and expected output were precisely specified, whether all required input files were complete and accessible in the execution environment, and whether the expected output was compatible with the corresponding evaluator. They also inspected each 100-task stream for substantial anomalies in difficulty progression, while accounting for both task difficulty and the intended skill-dependency structure. No major executability issue or sequence-level ordering inconsistency was identified; the remaining variations were localized and did not disrupt the overall progression.

To assess the reliability of the LLM-based dependency evaluation, the same annotators independently labeled 400 sampled directional judgments---200 each from Finance and Law---using the three-label rubric described above and without access to the LLM judge's predictions. Disagreements were resolved through discussion, and the resulting adjudicated labels were used to compute human--LLM agreement.

The adjudicated human labels exactly matched the GPT-5.4 judgments in 190 of 200 Finance cases (95.0\%) and 166 of 200 Law cases (83.0\%), yielding an overall agreement of 356/400 (89.0\%). Table~\ref{tab:dependency_human_validation} reports the label distribution and domain-level agreement.

\section{Structural Validation of Skill Continuity}
\label{app:skill_continuity}

This section provides the definitions and domain-level results for the structural validation introduced in
Section~\ref{subsec:skill_continuity}. The purpose of this analysis is to examine whether tasks in the curated sequences require recurring core skills, particularly within a local history window.

\subsection{Task-level Skill Annotation}

We use a locally served \texttt{Qwen3-32B} model as the skill annotator. The model is deployed through vLLM with an
OpenAI-compatible interface. We use temperature 0 and random seed 42, and disable the model's thinking mode. The same model configuration and annotation prompt are used for all five domains.

Each task is annotated independently. The annotator receives the task instruction, any workspace-file metadata, and up to five evaluation requirements when available. It does not receive the task's sequence position, source benchmark, reference answer, or annotations from other tasks. In particular, the sequence index is removed before the annotation request is constructed, preventing the intended order from directly influencing the extracted skills.

The annotation prompt asks the model to identify the smallest set of reusable capabilities or procedures needed to complete the task. A valid skill must describe an observable operation that could be reused in another task, such as \textit{table information extraction}, \textit{percentage-change calculation}, or \textit{statutory-rule application}, rather than a broad topic such as ``finance'' or a task-specific description such as ``solve this problem.'' For each task, the model returns between one and eight skills with the following fields:

\begin{itemize}
    \item a lowercase, snake-case skill name;
    \item a short description of the reusable capability;
    \item a skill type from \textit{knowledge}, \textit{reasoning},
    \textit{tool}, \textit{workflow}, or \textit{output};
    \item an importance label of either \textit{core} or
    \textit{support}; and
    \item a short task-grounded justification.
\end{itemize}

Core skills are the main capabilities exercised by the task, whereas supporting skills describe secondary operations that assist task completion. We apply deterministic post-processing to the model outputs. Skill names are converted to lowercase snake case, duplicate names within a task are removed, and malformed outputs are rejected and retried. Skills that describe only benchmark-interface requirements---such as multiple-choice selection, exact-match formatting, JSON answer wrappers, or mandated output filenames---are removed. Retrieval is retained as a skill only when the task requires locating information outside the supplied materials. These rules prevent superficial similarities in evaluation format from being counted as transferable task-solving skills.

For the structural analysis, we retain only skills labeled as \textit{core}. Let $\mathcal{S}_i$ denote the set of core skills required by the task at position $i$. The annotation process produces 318 unique core-skill names in Law, 335 in Finance, 345 in Healthcare, 262 in Mathematics, and 284 in Office. These are free-form skill annotations rather than labels selected from a predefined ontology; the semantic matching procedure described below is therefore used to account for differently named skills that express similar capabilities.

\begin{tcolorbox}[
    colback=gray!3,
    colframe=gray!50,
    sharp corners,
    boxrule=0.5pt,
    fontupper=\small,
    title={Task-level Skill Annotation Prompt (abridged)}
]
Given a task, identify the smallest reusable skills that an agent must
execute to solve it. A skill should describe an observable capability
or procedure rather than a broad topic or a description specific to
this task.

Return 1--8 skills. For each skill, provide a lowercase snake-case
name, a reusable description, its type
(\texttt{knowledge}, \texttt{reasoning}, \texttt{tool},
\texttt{workflow}, or \texttt{output}), its importance
(\texttt{core} or \texttt{support}), and a short task-grounded
justification.

Ignore benchmark-interface mechanics such as selecting an answer
option, exact-match formatting, JSON answer wrappers, and mandated
output filenames. Include retrieval only when the task requires
locating information outside the supplied materials.
\end{tcolorbox}

Because independently generated skill names may differ despite describing similar capabilities, exact string matching provides an incomplete estimate of skill reuse. We therefore construct a textual representation for each skill by combining its normalized name with up to two task-independent descriptions. We encode these representations using \texttt{sentence-transformers/all-mpnet-base-v2} and compute their cosine similarity. Two skills $s$ and $s'$ are treated as semantic counterparts when

\begin{equation}
    \operatorname{sim}(s,s') \geq \tau,
    \qquad \tau = 0.85.
\end{equation}

Exact normalized-name matches are included as semantic matches. The primary analysis uses $\tau=0.85$, while results under alternative thresholds are reported below.

\subsection{Human Validation of Skill Annotations}
\label{app:skill_annotation_validation}

To evaluate the reliability of the \texttt{Qwen3-32B} skill annotations, we conduct a human validation study on 50 tasks stratified across the five domains, with 10 tasks sampled from each domain. For every target task, we construct a randomized, provenance-blinded candidate pool containing the Qwen-proposed core skills for that task and hard distractors sampled from the skill annotations of other tasks in the same domain. Using same-domain distractors makes the validation more challenging than distinguishing relevant skills from unrelated cross-domain concepts.

Two human annotators independently select the skills they consider necessary for completing each target task. They are not shown whether a candidate originates from the target-task annotation or from the distractor pool. Separately, they assess whether the proposed annotations provide at least mostly complete coverage of the core capabilities required by the task.

Table~\ref{tab:skill_annotation_human_validation} summarizes the selection results. The two annotators achieved 89.5\% option-level agreement, Cohen's $\kappa=0.790$, and a selected-set Jaccard similarity of 81.8\%. Averaged across annotators, 92.6\% of the Qwen-proposed skills are accepted, whereas only 5.2\% of the hard distractors are selected. Under the more conservative consensus criterion, both annotators accept 167 of the 196 proposed skills (85.2\%), but only 4 of the 165 distractors (2.4\%). Consequently, 97.7\% of the options jointly selected as core skills originate from the Qwen annotations. No Qwen-proposed skill is rejected by both annotators, and both annotators judge 49 of the 50 tasks to have at least mostly complete skill coverage.

\begin{table}[t]
\centering
\caption{Human validation of task-level core-skill annotations. Mean
selection averages the two annotators; consensus requires both
annotators to select the skill.}
\label{tab:skill_annotation_human_validation}
\small
\setlength{\tabcolsep}{6pt}
\begin{tabular}{@{}lcc@{}}
\toprule
\textbf{Metric}
& \textbf{Proposed}
& \textbf{Distractors} \\
\midrule
Candidate options    & 196          & 165        \\
Mean selection       & 92.6\%       & 5.2\%      \\
Consensus selection  & 167 (85.2\%) & 4 (2.4\%)  \\
\bottomrule
\end{tabular}
\end{table}

These results indicate that the automatically generated annotations capture task-relevant core capabilities with high human acceptance and coverage, while rarely selecting plausible same-domain distractors. They therefore support the use of these annotations in the subsequent skill-continuity analysis.

\subsection{Skill-continuity Metrics}

For a task at position $i$, we consider a history window containing the previous $w$ tasks,

\begin{equation}
    \mathcal{H}^{(w)}_i =
    \left\{
        \mathcal{S}_j
        \mid
        \max(1,i-w) \leq j < i
    \right\}.
\end{equation}

When the complete preceding history is used, we write $\mathcal{H}^{(\mathrm{all})}_i$. A core skill $s\in\mathcal{S}_i$ is considered covered if at least one skill in the selected history is its semantic counterpart. The historical core-skill coverage of task $i$ is therefore

\begin{equation}
    C^{(w)}_i =
    \frac{1}{|\mathcal{S}_i|}
    \sum_{s\in\mathcal{S}_i}
    \mathbbm{1}
    \left[
        \max_{\substack{
            s' \in \mathcal{S}_j\\
            \max(1,i-w)\leq j<i
        }}
        \operatorname{sim}(s,s')
        \geq \tau
    \right].
\end{equation}

We summarize the sequence using two metrics:

\begin{itemize}
    \item \textbf{Task reuse rate}: the fraction of tasks after the
    first for which $C^{(w)}_i>0$, meaning that at least one required
    core skill has a counterpart in the selected history.

    \item \textbf{Mean core-skill coverage}: the average value of
    $C^{(w)}_i$ over all tasks after the first. This measures the
    fraction of each target task's core-skill requirements that have
    appeared previously.
\end{itemize}

All reported cross-domain averages are macro-averages: we first calculate each metric separately within every domain and then average the five domain-level values.

\subsection{Overall Skill Recurrence}

Table~\ref{tab:skill_continuity_domain} reports the results when the complete preceding history is used. Across domains, between 63.6\% and 77.8\% of eligible tasks reuse at least one previously encountered core skill. The corresponding mean core-skill coverage ranges from 23.2\% in Healthcare to 46.2\% in Office. Macro-averaged across the five domains, the task reuse rate is 69.5\%, and the mean core-skill coverage is 35.5\%.

\begin{table}[t]
\centering
\caption{Core-skill recurrence within the preceding task history at cosine threshold $\tau=0.85$. The task reuse rate is the percentage of tasks after the first with at least one matched core skill. Mean coverage is averaged over target tasks within each domain.}
\label{tab:skill_continuity_domain}
\small
\setlength{\tabcolsep}{5pt}
\begin{tabular}{lcc}
\toprule
\textbf{Domain}
& \textbf{Task reuse rate}
& \textbf{Mean coverage} \\
\midrule
Law        & 67.7 & 29.8 \\
Finance    & 67.7 & 36.6 \\
Healthcare & 63.6 & 23.2 \\
Mathematics& 70.7 & 41.6 \\
Office     & 77.8 & 46.2 \\
\midrule
\textbf{Macro average}
& \textbf{69.5}
& \textbf{35.5} \\
\bottomrule
\end{tabular}
\end{table}

\subsection{Comparison with Random Task Orders}

Overall recurrence alone does not determine whether related skills occur near one another. We therefore compare the curated order with random permutations using local history windows of $w\in\{1,5,10\}$. For every domain and window size, we randomly permute the 100 tasks 10,000 times while keeping the task annotations fixed. For each permutation, we recompute the mean core-skill coverage using the same history window.

Let $C_{\mathrm{obs}}^{(w)}$ denote the coverage of the curated order and let $C_b^{(w)}$ denote the coverage of random permutation $b$. We report the ordered-minus-random difference

\begin{equation}
    \Delta^{(w)}
    =
    C_{\mathrm{obs}}^{(w)}
    -
    \frac{1}{B}\sum_{b=1}^{B} C_b^{(w)},
    \qquad B=10{,}000.
\end{equation}

The one-sided permutation-test $p$-value is calculated as

\begin{equation}
    p =
    \frac{
        1 + \sum_{b=1}^{B}
        \mathbbm{1}
        [C_b^{(w)} \geq C_{\mathrm{obs}}^{(w)}]
    }{B+1}.
\end{equation}

\begin{figure*}[t]
    \centering
    \includegraphics[width=0.92\textwidth]
    {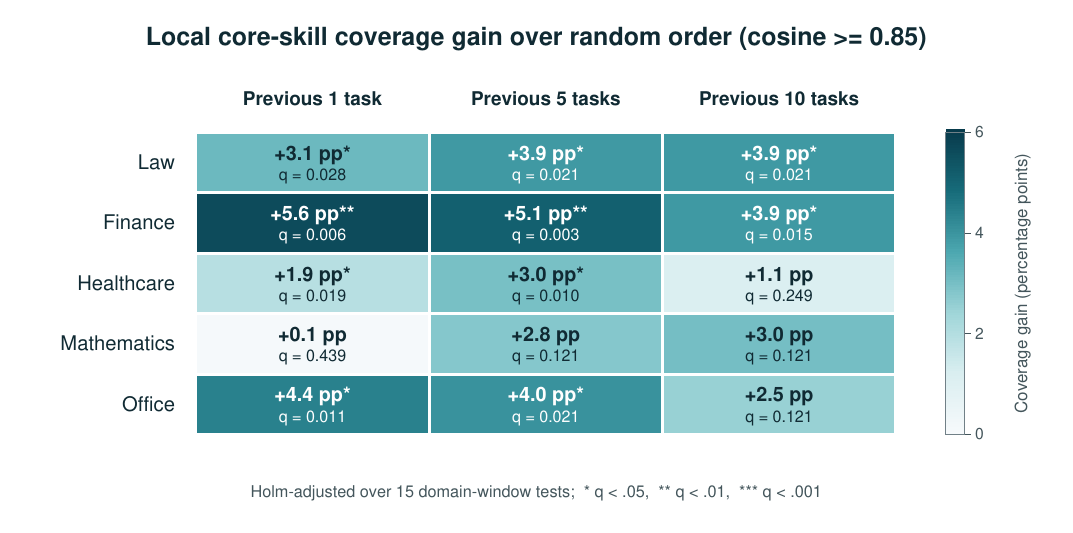}
    \caption{
    Difference in mean core-skill coverage between the curated sequences and random task permutations. Positive values indicate that the curated order places semantically related core skills closer together than expected under random ordering. The local windows cover the preceding 1, 5, or 10 tasks. Reported $q$-values are Holm-adjusted; ${}^*q<0.05$, ${}^{**}q<0.01$, and ${}^{***}q<0.001$.
    }
    \label{fig:skill_continuity_windows}
\end{figure*}

We apply Holm correction jointly over the displayed domain--window comparisons. Figure~\ref{fig:skill_continuity_windows} shows the resulting coverage differences.

The curated order achieves higher coverage in all five domains and all three local windows. The improvement is positive in all 15 domain--window comparisons and remains significant after Holm correction in ten comparisons. The largest gains occur in Finance, where coverage exceeds the random-order mean by 5.6, 5.1, and 3.9 percentage points for windows of 1, 5, and 10 tasks, respectively. Law shows significant improvements across all three windows, while Healthcare and Office show significant improvements in the shorter local windows. Mathematics exhibits positive differences for all three windows, although these differences do not remain significant after correction.

These results indicate that the ordered sequences possess a clear local skill-continuity structure: tasks appearing near one another are more likely to require overlapping or semantically related core skills than under random ordering. The analysis establishes structural opportunities for skill reuse; whether agents successfully exploit these opportunities is evaluated separately through the sequential experiments.

\subsection{Sensitivity to the Similarity Threshold}

Table~\ref{tab:skill_threshold_sensitivity} reports the macro-averaged results under exact normalized-name matching and three semantic similarity thresholds. The central conclusion is stable across the tested thresholds. Even with the stricter threshold of 0.90, 63.8\% of tasks reuse at least one previous core skill. Exact matching gives a more conservative lower bound, as it does not merge differently named descriptions of similar capabilities.

\begin{table}[t]
\centering
\caption{Sensitivity of the all-history recurrence results to the
semantic similarity threshold. Exact denotes normalized-name matching
without semantic extension.}
\label{tab:skill_threshold_sensitivity}
\small
\setlength{\tabcolsep}{6pt}
\begin{tabular}{lcc}
\toprule
\textbf{Matching rule}
& \textbf{Task reuse rate}
& \textbf{Mean coverage} \\
\midrule
Exact name       & 59.6 & 26.7 \\
Cosine $\geq0.80$& 79.4 & 44.5 \\
Cosine $\geq0.85$& 69.5 & 35.5 \\
Cosine $\geq0.90$& 63.8 & 29.4 \\
\bottomrule
\end{tabular}
\end{table}

\subsection{Robustness to LLM-Based Functional Equivalence}
\label{app:gemma_skill_matching}

The primary analysis identifies semantically corresponding skills using embedding similarity. Although this approach is reproducible and computationally efficient, lexical similarity may not always reflect functional equivalence: differently worded skills may describe the same operation, while similarly worded skills may require distinct procedures. We therefore repeat the local skill-continuity analysis using \texttt{Gemma-3-12B} as an independent functional-equivalence judge.

\paragraph{Functional-equivalence judgments.}
We evaluate all 240,065 within-domain unordered pairs of Qwen-annotated core skills. For each pair, Gemma receives only the normalized skill names and their task-independent descriptions. It does not receive the embedding scores, task identities, sequence positions, or labels produced by the embedding-based matcher. Gemma assigns each pair a continuous functional-equivalence score between 0 and 1, with larger values indicating that the two skills describe interchangeable or strongly transferable operations. We use a strict threshold of 0.90 to construct the Gemma-based equivalence graph.

The judge prompt is shown below.

\begin{tcolorbox}[
    colback=gray!3,
    colframe=gray!50,
    sharp corners,
    boxrule=0.5pt,
    fontupper=\ttfamily\small,
    title={Functional-Equivalence Judge Prompt}
]
You are evaluating whether two skills required by AI-agent tasks are
functionally equivalent.

\vspace{\baselineskip}
\textbf{Skill A}

Name: \{skill\_a\_name\}\\
Description: \{skill\_a\_description\}

\vspace{\baselineskip}
\textbf{Skill B}

Name: \{skill\_b\_name\}\\
Description: \{skill\_b\_description\}

\vspace{\baselineskip}
\textbf{Instruction}

Judge the extent to which the two skills represent the same reusable
operation or capability. Focus on what an agent must actually do,
rather than lexical overlap between the names. Skills may be
functionally equivalent even when they are described using different
words. Conversely, similarly named skills may be distinct if they
require different operations.

Assign a continuous score from 0 to 1, where 1 indicates functional
equivalence and 0 indicates unrelated capabilities.

\vspace{\baselineskip}
\textbf{Required output}

Return a JSON object with the following fields:

\{
  "score": <number between 0 and 1>,\\
  "reason": "<brief explanation>"
\}
\end{tcolorbox}

\paragraph{Comparison with embedding-based matching.}
The continuous Gemma and embedding scores are moderately correlated (Pearson $r=0.553$; Spearman $\rho=0.516$), but their thresholded equivalence graphs differ substantially. At their respective thresholds, Gemma identifies 876 equivalent skill pairs, whereas the embedding-based matcher identifies 249. Their positive-pair Jaccard overlap is only 0.180.

Qualitative disagreements illustrate the limitations of lexical matching. For example, \texttt{cross\_document\_reconciliation} and \texttt{multi\_document\_evidence\_synthesis} receive an embedding similarity of 0.589 but a Gemma score of 0.95, indicating a functionally related pair expressed using different terminology. Conversely, \texttt{json\_file\_reading} and \texttt{jsonl\_file\_reading} receive an embedding similarity of 0.930 but a Gemma score of 0.30, reflecting operational differences that are obscured by their lexical similarity.

\paragraph{Skill-continuity results.}
Using the Gemma-based equivalence graph, we recompute mean core-skill coverage for a five-task history window. As in the primary analysis, we compare the curated sequence against 10,000 random task permutations independently within each domain and apply Holm correction across the five comparisons.

Table~\ref{tab:gemma_skill_continuity} reports the results. The curated sequences achieve higher local skill coverage than random ordering in all five domains. Macro-averaged coverage increases from 20.4\% under random permutations to 25.7\% under the curated order, an improvement of 5.3 percentage points. Every domain remains significant after Holm correction.

\begin{table}[t]
\centering
\caption{Five-task local skill continuity under independent
\texttt{Gemma-3-12B} functional-equivalence judgments. The random
baseline is averaged over 10,000 task-order permutations per domain.}
\label{tab:gemma_skill_continuity}
\small
\setlength{\tabcolsep}{4.5pt}
\begin{tabular}{lrrrr}
\toprule
\textbf{Domain}
& \textbf{Curated}
& \textbf{Random}
& \textbf{Gain}
& \textbf{$p_{\mathrm{Holm}}$} \\
\midrule
Law        & 21.6 & 15.7 & +5.9 & 0.0006 \\
Finance    & 26.2 & 21.2 & +5.0 & 0.0005 \\
Healthcare & 16.5 & 13.5 & +3.0 & 0.0061 \\
Mathematics& 31.1 & 24.7 & +6.4 & 0.0002 \\
Office     & 33.4 & 27.2 & +6.2 & $<0.0001$ \\
\midrule
\textbf{Macro average}
           & \textbf{25.7}
           & \textbf{20.4}
           & \textbf{+5.3}
           & -- \\
\bottomrule
\end{tabular}
\end{table}

Despite producing a substantially different equivalence graph, the independent Gemma judge recovers the same benchmark-level conclusion as the embedding-based analysis: the curated sequences place functionally related skills closer together than expected under random ordering. The observed local skill-continuity structure is therefore not an artifact of cosine similarity or the selected embedding threshold.

\section{Full Prompts for the Three-Turn Interaction Protocol}
\label{app:prompts}

Each subtask in Sequential execution follows the same three-turn
interaction protocol:

\begin{enumerate}
    \item \textbf{Task Introduction.}
    The agent receives the subtask instruction together with its
    current skill repository. It is prompted to review the available
    skills and determine whether any are relevant to the task.

    \item \textbf{Task Execution.}
    The agent performs the task using the available skills, tools, and
    external resources when necessary. Its actions and outputs are
    recorded for evaluation.

    \item \textbf{Reflection and Skill Update.}
    After execution, the agent receives evaluator feedback, such as
    programmatic test results or rubric-level scores. It is prompted to
    diagnose failures and determine whether the experience should be
    consolidated into a reusable skill. The agent may create a new
    skill or revise an existing one using the \textbf{Create Skill} and
    \textbf{Modify Skill} meta-skills. The updated repository is
    available in subsequent subtasks.
\end{enumerate}

\begin{figure}[ht]
    \centering
    \includegraphics[width=\columnwidth]
    {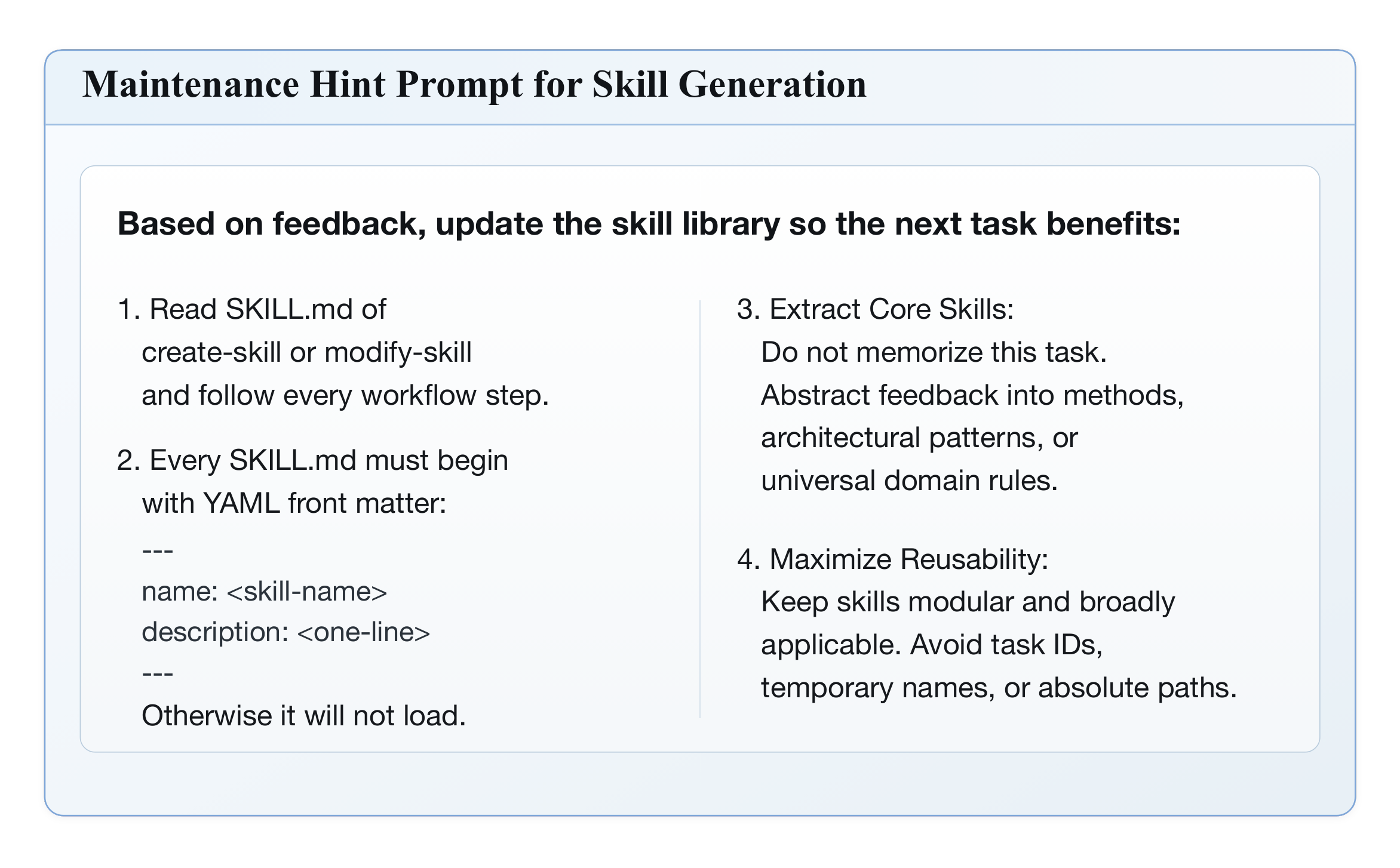}
    \caption{Prompt used during the reflection and skill-maintenance
    turn. The agent is instructed to create or revise valid skills
    through the \textbf{Create Skill} and \textbf{Modify Skill}
    meta-skills.}
    \label{fig:skillprompt}
\end{figure}

Figure~\ref{fig:skillprompt} shows the instruction used during the
skill-maintenance turn. The complete prompts for all three turns are
provided below.

\begin{tcolorbox}[
    colback=gray!3,
    colframe=gray!50,
    sharp corners,
    coltitle=black,
    fonttitle=\bfseries,
    boxrule=0.5pt,
    left=6pt,
    right=6pt,
    top=6pt,
    bottom=6pt,
    fontupper=\ttfamily\small,
    title={Sequential Agent Prompt: Run Orientation}
]
\textbf{Turn 0: Sequential run orientation (injected once before all subtasks).}

You are running inside Harbor's Codex Sequential Agent.

This session will present \{n\_tasks\} task(s) one at a time. For each task you will receive three turns:

\quad Turn 1 --- Task + skill index: the task instruction plus a listing of available skills.\\
\quad Turn 2 --- Execute: complete the task.\\
\quad Turn 3 --- Judge feedback + skill maintenance: receive an automated judge score and feedback, then update the skill library using the create-skill and modify-skill meta-skills so future tasks benefit.

Rules:\\
- Prefer skills found under \{skills\_dir\}/. Inspect them before inventing a new approach.\\
- After completing each task, the judge will run automatically --- wait for Turn 3.\\
- Use create-skill to add a new skill and modify-skill to update an existing one. Both meta-skills live in \{skills\_dir\}.
\end{tcolorbox}

\begin{tcolorbox}[
    colback=gray!3,
    colframe=gray!50,
    sharp corners,
    coltitle=black,
    fonttitle=\bfseries,
    boxrule=0.5pt,
    left=6pt,
    right=6pt,
    top=6pt,
    bottom=6pt,
    fontupper=\small
]
\textbf{Turn 1: Task instruction.}

\texttt{[Task \{task\_num\} of \{n\_tasks\}]}\\
\texttt{\{sub\_instruction\}}

\vspace{\baselineskip}
\textbf{If this is not the first subtask, the following memory hint is prepended:}

Previous task context and judge feedback are at
\texttt{/root/task\_memory.md}; read it before starting.

\texttt{[Task \{task\_num\} of \{n\_tasks\}]}\\
\texttt{\{sub\_instruction\}}

\textbf{Turn 2: Execution.}

No additional user prompt is injected in this turn. The Codex agent executes the current subtask after receiving the Turn 1 instruction. After execution, Harbor runs the automated judge externally and writes feedback to
\texttt{/logs/verifier/task\_\{task\_num\}\_feedback.json}.

\textbf{Turn 3: Judge feedback and skill maintenance.}

Judge feedback for task \texttt{\{task\_num\}} is now available at
\texttt{/logs/verifier/task\_\{task\_num\}\_feedback.json}. Read the file and review the score and rubric breakdown.

\vspace{0.5\baselineskip}
Based on this feedback, update the skill library so the next task benefits:

\begin{enumerate}
    \item Read the \texttt{SKILL.md} of \texttt{create-skill} or \texttt{modify-skill} in \texttt{\$HOME/.agents/skills/}, and follow its workflow exactly.
    \item Every \texttt{SKILL.md} you write or modify must begin with YAML front matter.
    Skills without the header will not be loaded by the framework.
    \item Extract core skills: do not memorize the solution or hardcode fixes for this specific task. Abstract feedback into a reusable methodology, architectural pattern, or universal domain rule.
    \item Maximize reusability: keep the documented skill modular and broadly applicable to future unseen tasks. Avoid hardcoding task IDs, temporary variable names, or absolute file paths.
\end{enumerate}
\end{tcolorbox}

\section{Example Task Format and Skill-Dependency Chain}

To make the benchmark format more concrete, we show an example chain from the
\textbf{Finance} domain. We omit irrelevant tables, filings, and data fields for space.

\begin{tcolorbox}[colback=gray!3,
    colframe=gray!50,
    sharp corners,
    coltitle=black,
    fonttitle=\bfseries,
    boxrule=0.5pt,
    left=6pt,
    right=6pt,
    top=6pt,
    bottom=6pt,
    fontupper=\small
]
Task 1: ... What is the amount of cash and cash equivalents held by the company in 2018 and 2019 respectively?

\vspace{\baselineskip}
Task 7: ... What was the amount of value-added tax receivables, net, noncurrent in 2019?

\vspace{\baselineskip}
Task 35: ... What was the percent growth of The Priceline Group Inc. from 2014 to 2015?

\vspace{\baselineskip}
Task 50: Does 3M have a reasonably healthy liquidity profile based on its quick ratio for Q2 FY2023? If the quick ratio is not relevant to measuring liquidity, state that and explain why.

\vspace{\baselineskip}
Task 81: ... Requirements:

1. Read the provided financial data files.

2. Calculate the required financial components.

3. Compute the weights for equity and debt in the capital structure.

4. Calculate the final WACC using the given formula.

5. Handle edge cases.
\end{tcolorbox}

This chain illustrates how reusable skills can transfer across subtasks. Early tasks such as Task 1 and Task 7 require locating values in financial tables and matching rows, columns, periods, and units. Later tasks reuse these operations as subroutines for more complex reasoning: Task 35 adds percentage-growth calculation, Task 50 adds ratio computation and financial interpretation, and Task 81 requires a programmatic WACC workflow with data loading, formula application, and output
validation. Thus, skills for table parsing, period-specific value extraction, unit normalization, formula application, and validation can be learned early and reused in later subtasks.

\section{Additional Ablation: Retrieval-Augmented Trajectory Memory}
\label{app:rag_ablation}

We also evaluate a retrieval-augmented generation (RAG) baseline, where prior trajectory snippets are indexed and retrieved for later subtasks instead of maintaining an explicit skill library. The results are shown in Table~\ref{tab:rag_seq_results}.

\begin{table}[t]
\centering
\caption{RAG ablation on Opus 4.7. RAG retrieves prior trajectory context, while Seq. maintains explicit skills.}
\label{tab:rag_seq_results}
\resizebox{\columnwidth}{!}{
\begin{tabular}{ll ccc cc}
\toprule
\multirow{2}{*}{\textbf{Domain}} &
\multirow{2}{*}{\textbf{Setting}} &
\multicolumn{3}{c}{\textbf{Subtask Rewards}} &
\multicolumn{2}{c}{\textbf{Reward}} \\
\cmidrule(lr){3-5} \cmidrule(lr){6-7}
& & \textbf{EM} & \textbf{Prog.} & \textbf{Rubric} & \textbf{Raw} & \textbf{Norm.} \\
\midrule

\multirow{2}{*}{\textbf{Healthcare}}
& RAG  & 0.815 & 0.500 & \textbf{0.539} & 0.574 & 0.617 \\
& Seq. & \textbf{0.926} & 0.500 & 0.517 & \textbf{0.631} & \textbf{0.635} \\

\bottomrule
\end{tabular}}
\end{table}

The RAG baseline shows a similar pattern to the ICL setting. It improves
rubric-judged subtasks, which suggests that part of the rubric gain may be a
side effect of adapting to prior feedback and response patterns rather than
evidence of reusable skill acquisition. In contrast, Seq. remains stronger on
exact-match rewards, indicating that explicit skill maintenance is more useful
for enforcing strict output formats, answer normalization, and step-by-step
verification. This echoes the conclusion in Section~\ref{sec:icl_ablation}:
contextual adaptation can help open-ended rubric-style answers, while explicit
skills provide a more disciplined workflow for tasks with stricter correctness
constraints.

\section{Analysis on Generated Skills}
\label{app:generated skills}
We further analyze the quality of generated skills using GPT-4.1-mini. All generated skills are extracted and sent to the LLM through the following prompt: 
\begin{tcolorbox}[colback=gray!3,
    colframe=gray!50,
    sharp corners,
    coltitle=black,
    fonttitle=\bfseries,
    boxrule=0.5pt,
    left=6pt,
    right=6pt,
    top=6pt,
    bottom=6pt,
    fontupper=\small
]
You are a strict evaluator of agent Skill files.

A Skill is a reusable instruction file that an agent may load before future
tasks. Good Skills are not task memorization: they abstract a reusable method,
say when to use it, provide executable steps or formulas/templates, include
pitfalls/validation checks, and stay concise enough to be useful.

Judge only the provided Skill content. Do not reward domain correctness beyond
what can be inferred from the text. Penalize missing/truncated content, raw
placeholders, task-specific paths, or instructions that would not generalize.
Return only valid JSON matching the requested schema.

Score the requested indicators directly. Do not invent an overall score; it
will be computed outside the judge.

\vspace{\baselineskip}
Format/loadability indicators, each 0 or 1:

- front\_matter\_valid: has a valid YAML front matter block.

- name\_present: YAML front matter contains a non-empty name field.

- name\_match: YAML name matches the skill name/title.

- description\_present: has a one-sentence description.

\vspace{\baselineskip}
Content-quality indicators, each 0, 0.5, or 1:

- specificity: clear use case, trigger, or scope boundary.

- procedure\_score: executable steps, formulas, templates, checklist, or decision tree.

- generalization\_score: abstracts a reusable method rather than memorizing one task.

- anti\_pitfall\_score: includes common mistakes, counterexamples, edge cases, or failure modes.

- verification\_score: includes validation/checklist/gate before final output.

- minimality: neither too short to be useful nor bloated/distracting.

- toolability: contains reusable code, command, schema, formula, table, or template.

- domain\_grounding: includes necessary domain-specific safety/compliance/source constraints.

\end{tcolorbox}
We compute the final skill-quality score as
\[
\text{score} = \text{format} \times \text{content}.
\]
A skill with high-quality content but invalid formatting receives a score of 0, because it cannot be reliably loaded and reused by the agent in future subtasks. The results are shown in Table~\ref{tab:skill_quality_eval}.

\begin{table}[t]
\centering
\caption{Quality evaluation of generated skills across models and domains. Skill pool size here counts only agent-generated skills, excluding meta skills and initially provided skills.}
\label{tab:skill_quality_eval}
\resizebox{0.9\columnwidth}{!}{
\begin{tabular}{llcc}
\toprule
\textbf{Model} & \textbf{Domain} & \textbf{Skill Pool Size} & \textbf{Avg. Quality Score} \\
\midrule
GPT-4o & Finance & 48 & 5.79 \\
GPT-5.3-Codex & Finance & 28 & 7.89 \\
\midrule
GPT-4o & Healthcare & 81 & 5.36 \\
GPT-5.3-Codex & Healthcare & 38 & 8.19 \\
\midrule
GPT-4o & Law & 72 & 5.25 \\
GPT-5.3-Codex & Law & 43 & 7.95 \\
\midrule
GPT-4o & Math & 97 & 5.74 \\
GPT-5.3-Codex & Math & 48 & 7.43 \\
\midrule
GPT-4o & Office & 86 & 6.24 \\
GPT-5.3-Codex & Office & 48 & 8.27 \\
\midrule
GPT-4o & Overall & 384 & 5.68 \\
GPT-5.3-Codex & Overall & 205 & 7.94 \\
\bottomrule
\end{tabular}}
\end{table}
These results provide additional evidence for the skill-fragmentation pattern discussed in Finding~\ref{subsec:skill_dynamics}. GPT-4o produces a much
larger number of skills across domains (384 in total), but their average quality score is substantially lower (5.68). In contrast, GPT-5.3-Codex maintains a smaller skill pool (205 skills) with consistently higher quality (7.94 on average). This suggests that stronger models are better at consolidating repeated experience into compact and reusable abstractions, whereas weaker models tend to generate more fragmented or task-specific skills that are less useful for future subtasks. Thus, the quality evaluation supports our interpretation that skill fragmentation is one reason why weaker models benefit less reliably from long-term sequential skill learning.

\section{Examples of Generated Skills}
To illustrate what generated skills look like, we present one example from each of GPT-4o, GPT-5.3-Codex, and Claude 4.7 Opus.
\begin{skillbox}
This skill was generated by GPT-4o.
---
name: date-range-filter
description: Filter data by specific date range for trading tasks.
---

# Date Range Filter

## What it does
Filters trading data within a specified start date to one day before the end date.

## Implementation
```python
def filter_dates(data, start_date, end_date):
    filtered_data = {}
    for date, entry in data.items():
        if start_date <= date < end_date:
            filtered_data[date] = entry
    return filtered_data

# Example Usage
filtered_data = filter_dates(data, 2020-10-01, 2021-05-05)
```

## Key Parameters / Notes
- `start_date` and `end_date` should be in YYYY-MM-DD format.
- Excludes `end_date` itself in filtering.
- Ensures compliance with trading period requirements.
\end{skillbox}

\begin{skillbox}
This skill was generated by GPT-5.3-Codex.
---
name: tatqa\-metric\-alignment
description: Align financial QA answers to the exact asked metric by reconciling table rows, subtotals, and narrative context with unit and period checks.
---

# TAT-QA Metric Alignment

## What It Does
Prevents extraction mistakes in financial table QA by forcing a metric-match check before selecting values.

## Workflow
1. Parse the question target exactly:
- Metric noun phrase (for example: `cash and cash equivalents`, `total liquidity`, `current assets`).
- Requested periods and order (for example: `2018 and 2019 respectively` means output `2018 | 2019`).
- Required granularity (`exact table amount` vs `approximately` phrasing).

2. Build candidate values from both sources:
- Table row values (line items).
- Table subtotal/total rows.
- Narrative context values (especially management summary sentences).

3. Resolve source selection with this priority:
- If the question asks for a line item and only the table has that line item, use the line item.
- If the question wording is broad (`had`, `liquidity`, `available`) and context explicitly states a combined figure, prefer the context-stated combined amount.
- If context restates table values at different precision (for example thousands vs millions), match the style used in the wording expected by the prompt/judge.

4. Validate before finalizing:
- Unit check: thousands vs millions.
- Composition check: line item vs subtotal vs combined categories.
- Time-order check: ensure output order matches `respectively` order.

## Common Pitfalls
- Returning a component (`cash & cash equivalents`) when the prompt/judge expects a combined liquidity figure (`cash + short-term investments`).
- Mixing unit scales (`18,304` in thousands vs `\$18.3 million`).
- Reversing years when `respectively` is present.

## Output Rule Template
When format is `"$X | $Y"`:
- Map X to first requested period.
- Map Y to second requested period.
- Keep currency symbols and unit style consistent with selected source.
\end{skillbox}

\begin{skillbox}
This skill was generated by Claude 4.7 Opus.
---
name: medical-advice-openended
description: Answer open-ended medical advice questions (often in patient-facing language, possibly non-English) and produce a structured JSON output with findings, citations, and conclusions.
---

# Open-Ended Medical Advice Tasks

## Task Format
Some tasks present a patient question or scenario (sometimes in a non-English language: French, Spanish, etc.) and ask for clinically accurate guidance. Output is a JSON OBJECT (not a single letter) typically with keys like findings/citations/conclusions.

## CRITICAL: Always Write Output File First
Some turns are truncated. ALWAYS write `/app/task_{NNN}_output.json` BEFORE doing extensive exploration or reading. Write a first-pass JSON immediately, then refine.

## Required Output
- Path: `/app/task_{NNN}_output.json` (zero-pad N to 3 digits)
- A JSON OBJECT — typical keys: `key_findings`, `citations`, `conclusions`, `recommendation`
- Language: respond in the same language as the prompt where possible.

## Workflow
1. Identify task number; determine output path.
2. Translate the prompt mentally if non-English.
3. Draft a structured answer with:
   - `key_findings`: bullet list of medically relevant facts
   - `country_specific` (when relevant): differences across major countries (US/CDC, France/HAS, UK/NHS, Germany/STIKO, WHO)
   - `recommendation`: clear, safe clinical guidance
   - `citations`: list of authoritative sources (CDC, WHO, HAS, NHS, NICE, UpToDate, peer-reviewed)
   - `safety_notes`: caveats (consult local physician, individual factors)
4. WRITE THE FILE.

## Common Topics & Quick References

### Pertussis Booster (whooping cough / coqueluche)
- Adults: One Tdap dose; then Td/Tdap booster every 10 years
- Pregnancy: Tdap each pregnancy (27-36 weeks US, 16-32 UK, 20-36 France)
- Cocoon strategy: parents, grandparents, caregivers of infants <12 mo
- Healthcare workers: one dose if not received as adult
- Country specifics:
  - US (CDC/ACIP): Tdap once for adults 19+, repeat Td every 10y
  - France (HAS): rappel adulte à 25 ans dTcaPolio; rattrapage jusqu'à 39 ans
  - UK (NHS): pregnant women routinely; no universal adult booster
  - Germany (STIKO): one adult booster with next Td
  - Canada: one adult Tdap; pregnancy 27-32 wks
  - Australia: adult booster recommended esp. for infant contacts

### Influenza vaccine
- Annual, all adults; CDC recommends for all ≥6 mo

### COVID-19 boosters
- Updated annually; risk-based for high-risk groups

### HPV
- Up to age 26 routine, 27-45 shared decision-making

### Shingles (zoster)
- Shingrix 2 doses, adults ≥50

## Example
```python
import json
output = {
    "key_findings": ["..."],
    "country_specific_recommendations": {"US": "...", "France": "..."},
    "recommendation": "...",
    "citations": ["CDC ACIP 2024", "HAS calendrier vaccinal 2024"],
    "safety_notes": "Consult your physician for individual factors."
}
with open("/app/task_002_output.json", "w") as f:
    json.dump(output, f, ensure_ascii=False, indent=2)
```

## Pitfalls
- DO NOT skip writing the file even if you only have a partial answer.
- Open-ended != MCQ. Check task wording: "JSON object with your key findings, citations, and conclusions" -> structured object.
- Some prompts arrive in French/Spanish — answer in the prompt language when natural.
- The verifier may use an LLM judge with a rubric; cover multiple dimensions (recommendation, country variation, safety, citations).
\end{skillbox}

These examples show that stronger models tend to produce skills with more structured workflows, explicit validation steps, and clearer pitfall awareness, which helps explain their larger and more reliable sequential gains in domains such as Finance and Healthcare.

\section{Pilot Evaluation with an Open-Source Model}
\label{app:open_source_pilot}

We additionally piloted Qwen3-32B, served locally through vLLM, using the same Codex-based sequential harness on the Finance stream. The run was terminated after 32 subtasks and is therefore not included as a directly comparable benchmark result. Even within this partial trajectory, performance was frequently limited by unreliable base-agent execution rather than by task-solving or skill-transfer capability. Across 233 logged tool calls, the model issued 23 unsupported function calls and 10 calls with malformed arguments; 118 of its 208 shell commands exited with a nonzero status. These failures included repeated invalid tool calls, malformed shell or Python commands, and failure to create the required output files.

A representative example occurred on Task~30, a programmatically evaluated cryptocurrency-trading task. The model made 58 tool calls, 57 of which failed, repeatedly modifying and reissuing malformed Python commands. It ultimately failed to create the required output file and received a score of zero. Some other subtasks were also affected by external judge API failures, so we do not report an aggregate reward for this incomplete pilot. Nevertheless, the tool traces indicate that, for this model--harness combination, sequential performance would be dominated by base execution failures rather than meaningful variation in continual skill accumulation. We therefore restrict the primary comparison to models that can reliably complete the underlying agentic tasks.

\begin{tcolorbox}[
    colback=gray!3,
    colframe=gray!50,
    boxrule=0.5pt,
    coltitle=black,
    left=6pt,
    right=6pt,
    top=6pt,
    bottom=6pt,
    fonttitle=\bfseries
    title={Abridged Qwen3-32B trajectory on Finance Task 30},
    fontupper=\small
]
\textbf{Agent:}
\begin{verbatim}
exec_command:
python3 -c "... data = [
  {'date': k, 'price': v for k, v in data.items()
] ..."
\end{verbatim}

\textbf{Environment:}
\begin{verbatim}
SyntaxError: closing parenthesis ']'
does not match opening parenthesis '{'
\end{verbatim}

\textbf{Agent:}
Repeated the same malformed construction across
multiple subsequent tool calls.

\textbf{Programmatic evaluator:}
\begin{verbatim}
AssertionError:
Missing output: /app/task_030_output.json
\end{verbatim}
\end{tcolorbox}

\end{document}